\documentclass[a4paper,fleqn]{cas-dc}

\usepackage[authoryear,longnamesfirst]{natbib}
\usepackage{graphicx}
\usepackage{lipsum}
\usepackage{amsfonts}
\usepackage{amssymb}
\usepackage{amsthm}
\usepackage{epstopdf}
\usepackage{algorithm, algorithmic}
\usepackage{enumitem}
\usepackage{color}
\usepackage{afterpage}
\usepackage{xcolor}
\usepackage{tabularx}
\newtheorem{assumption}{Assumption}
\newtheorem{theorem}{Theorem}
\newtheorem{lemma}[theorem]{Lemma}
\usepackage{mathtools}
\usepackage{subcaption}
\usepackage{float}
\DeclarePairedDelimiter{\floor}{\lfloor}{\rfloor}
\DeclarePairedDelimiter{\ceil}{\lceil}{\rceil}

\def\tsc#1{\csdef{#1}{\textsc{\lowercase{#1}}\xspace}}
\tsc{WGM}
\tsc{QE}

\begin{document}
\let\WriteBookmarks\relax
\def\floatpagepagefraction{1}
\def\textpagefraction{.001}

\shorttitle{}    

\shortauthors{}  

\title [mode = title]{PINNs Failure Region Localization and Refinement through White-box Adversarial Attack}  



%






\affiliation[1]{organization={School of Mathematics, Harbin Institute of Technology},
            addressline={92 W Dazhi St.}, 
            city={Harbin},
            postcode={150001}, 
            country={China}}
\cortext[1]{Corresponding author}

\affiliation[2]{organization={School of Astronautics, Harbin Institute of Technology},
            addressline={92 W Dazhi St.}, 
            city={Harbin},
            postcode={150001}, 
            country={China}}

\author[1]{Shengzhu Shi}
\author[1]{Yao Li}[orcid=0000-0002-1754-4528]
\cormark[1]
\ead{yaoli0508@hit.edu.cn}

\author[1]{Zhichang Guo}
\author[1]{Boying Wu}
\author[2]{Yang Zhao}









\begin{abstract}
Physics-informed neural networks (PINNs) have shown great promise in solving partial differential equations (PDEs). However, vanilla PINNs often face challenges when solving complex PDEs, especially those involving multi-scale behaviors or solutions with sharp or oscillatory characteristics. To precisely and adaptively locate the critical regions that fail in the solving process we propose a sampling strategy grounded in white-box adversarial attacks, referred to as “WbAR”. WbAR search for failure regions in the direction of the loss gradient, thus directly locating the most critical positions. WbAR generates adversarial samples in a random walk manner and iteratively refines PINNs to guide the model's focus towards dynamically updated critical regions during training. We implement WbAR to the elliptic equation with multi-scale coefficients, Poisson equation with multi-peak solutions, high-dimensional Poisson equations, and Burgers’ equation with sharp solutions. The results demonstrate that WbAR can effectively locate and reduce failure regions. Moreover, WbAR is suitable for solving complex PDEs, since locating failure regions through adversarial attacks is independent of the size of failure regions or the complexity of the distribution.  Code link: https://github.com/yaoli90/WbAR 
\end{abstract}


\begin{highlights}
\item We propose a white-box adversarial attack refinement strategy, namely WbAR, to accurately locate failure regions and adaptively refine PINNs.
\item We prove the adversarial attack is more beneficial than pure random walk in locating high-residual regions and give an upper bound for the generalization error of the model under WbAR. 
\item We implement WbAR to solve several challenging PDEs encountered by vanilla PINNs and demonstrate its efficiency in capturing irregular spatial characteristics, including multi-scale behavior, sharp or oscillatory solutions.
\end{highlights}



\maketitle

\section{Introduction}
Physics-informed neural networks (PINNs) represent a category of model-data hybrid-driven methods for solving partial differential equations (PDEs). 
They integrate physical laws and data measurements into the loss function, utilizing automatic differentiation to facilitate the learning process \cite{RPK19}.
PINNs offer numerous advantages, including: (i) being mesh-free, allowing them to potentially solve complex domain and high-dimensional problems; (ii) the flexibility to impose physical constraints and incorporate data information; (iii) the capability to handle nonlinear problems, supported by the universal approximation theorem and a variety of activation functions.

As a result, PINNs have emerged as promising paradigms for solving PDEs (forward problems) and estimating model parameters (inverse problems) in a broad spectrum of physical phenomena and applications. 
Examples include solving Euler equations with high-speed flows \cite{MJK20}, Navier-Stokes equations \cite{JCLK21}, phase-field equations \cite{MG22}, characterizing parameters in biological systems \cite{YLRK20}, and fluid dynamics \cite{RYK20}, among others. 
They are also gaining increasing attention in areas such as fractional equations \cite{PLK19}, integro-differential equations \cite{YNDH22}, and stochastic PDEs \cite{ZGK20}.

Despite obtaining fruitful results, vanilla PINNs face challenges when solving more complicated problems. 
To enhance their performance, numerous variants have been proposed based on the structure of neural networks and the properties of the PDEs. 
Some notable works include PINNs with new differentiation methods like CAN-PINN \cite{CWODO22}, hybrid-PINN \cite{F21}, new activation functions with adaptable hyper-parameters based on scaling factors \cite{JKK20}, and various temporal discretization methods like parallel PINN \cite{MKZK20}, bc-PINN \cite{MG22}, and causal-PINN \cite{wang2022respecting}. 
New loss designs, such as gPINNs \cite{YLMK22}, and formulations based on Galerkin methods like VPINNs \cite{VPINNs}, and hp-PINNs \cite{KZK21} have also been introduced. 
For a more comprehensive understanding of PINNs and their variants, we refer to the review papers \cite{LMMK21} \cite{CDGRRP22} \cite{KKLPWY21}.

While benefiting from the advantages of learning mechanisms, PINNs also suffer from the inherent limitations of deep neural networks. 
One of the most doubtful aspects is their robustness, i.e., a small perturbation to the input sample may lead to neural network misclassification or regression failure \cite{ning2023evaluating}. 
Therefore, attackers can manipulate the input data intentionally and cause the model to produce incorrect outputs. 
This type of attack is called an adversarial attack, and samples that cause the neural network to fail are referred to as adversarial samples. It has been studied that neural networks are generally not robust and are particularly vulnerable to adversarial attacks \cite{szegedy2013intriguing}.

One widely accepted explanation posits that neural networks locally exhibit linear properties with respect to the input \cite{goodfellow2014explaining}. 
It implies that small perturbations on each dimension can accumulate, leading to a significant variation in the loss due to the high dimensionality of the input.
In the context of PINNs applications, the input dimension is generally low, such that PINNs are robust against adversarial attacks across most of the sample space.
However, vulnerabilities to adversarial attacks may persist in regions characterized by multi-scale behaviors or solutions with sharp or oscillatory characteristics. 
This is attributed to the fact that in these regions, the loss gradient w.r.t. the input tends to be large because of the suddenly changed solution behaviors. Thus, by identifying adversarial samples of a PINN, one can locate the failure region and further use adversarial samples to improve the model's performance \cite{morris2020textattack}.
For more intuitive explanations, please refer to section \ref{sec:motivation}.

In this paper, we propose refining PINNs with white-box adversarial attacks refinement (WbAR), which fine-trains the model by adaptively adding new adversarial samples to improve model performance. 
For PINNs, adversarial samples can automatically and accurately adapt to the model failure locations and force the model to focus on the failure regions.
The failure location adaptation is independent from the volume of the failure region, the dimension of the problem or the distribution of the residual map, which is especially helpful in solving PDEs with specific characteristics such as multi-scale behavior, sharp or oscillatory solutions, etc.
\section{Related works and contributions}
\label{sec:related_works}
WbAR can be related to sampling strategies of PINNs, as it enhances PINNs through specially designed sampling. 
In the vanilla PINN algorithm, since PINNs are supposed to approximate the equation solution on the entire problem domain, Latin hypercube sampling (LHS) and uniform sampling are proposed to generate training samples \cite{RPK19}. While these sampling methods generally perform well for a wide range of cases, they may fail in solving more complex PDEs.


The unbiased sampling face challenges when the PDEs involving some irregular features such as multi-scale behavior, sharp or oscillatory solutions. 
To overcome these issues, various residual-based adaptive sampling methods have been proposed to enhance the performance of PINNs.
Lu et al. \cite{LMMK21} first proposed a residual-based adaptive refinement (RAR) method, which repeatedly adds new collocation points where the PDE residual is large. 
Mao et al. \cite{MM23} further divided the whole domain into several sub-domains and presented the adaptive sampling methods (ASMs) based on both the mean value of the residual for each sub-domain and the gradient of the solution. 
ASMs can eliminate the occurrence of bad residual points, resulting in a more stable iteration procedure. 
Different from RAR and ASMs, Wu et al. \cite{WZTKL23} resample new collocations by constructing a new probability density function with two hyperparameters based on the PDE residual. 
Another attractive work from the point view of probability density is the failure-informed PINNs (FI-PINNs), which use an effective failure probability based on the residual as the posterior error indicator to place more collocation points in the failure region \cite{gao2023failure}.
Two extensions of FI-PINN based on re-sampling technique and the subset simulation algorithm were consequently established in \cite{GTYZ23}.  

Both residual-based sampling strategies and WbAR aim to improve the performance of PINNs by adaptively adding new samples where the model fails and retraining on them. 
However, they differ in their approach to searching for these new samples. 
In the residual based methods, the search for high-residual regions is implemented through Monte Carlo sampling (e.g. RAR), brute force searching (e.g. ASMs), residual distribution approximating (e.g. FI-PINN), etc. 
Their effectiveness depends on the volume of the failure region, the dimension of the problem and the the distribution hypothesis of the residual map, respectively.    
On the other hand, WbAR locates failure regions directly through adversarial samples, independently of these factors. 

In fact, Shekarpaz et al. \cite{shekarpaz2022piat} mentioned applying PGD-based adversarial attacks and adversarial training, along with weight decay and Gaussian smoothing, to improve the performance of PINNs. 
However, the implementation details and the contribution of the adversarial attack were not carefully discussed.
The WbAR proposed in this paper provides a more efficient way for searching new samples where the model tends to fail. 
To illustrate the efficiency of WbAR, we first implement WbAR on a Poisson equation with multi-scale coefficients. 
WbAR can dynamically adjust the resolution to capture failures at each scale and produce satisfactory results where vanilla PINNs fail.
Next, for problems with multi-peak solutions and sharp solutions, we test WbAR on the Poisson equation and the Burgers' equation, respectively. 
WbAR can successfully identify and sample tiny or complicated distributed failure regions, resulting in precise results. Our experiments highlighted the effectiveness of the adversarial training in tackling various challenging problems encountered by PINNs.

The main contributions of this paper are threefold:
\begin{enumerate}
\item We propose a white-box adversarial attack refinement strategy, namely WbAR, to accurately locate failure regions and adaptively refine PINNs.
\item We prove the adversarial attack is more beneficial than pure random walk in locating high-residual regions and give an upper bound for the generalization error of the model under WbAR. 
\item We implement WbAR to solve several challenging PDEs encountered by vanilla PINNs and demonstrate its efficiency in capturing irregular spatial characteristics, including multi-scale behavior, sharp or oscillatory solutions.
\end{enumerate}

\subsection{Organization}
The rest of the paper is organized as follows: In section 2, we provide a brief review of PINNs and the concept of adversary attacks. In section 3, we propose the white-box adversarial attack refinement scheme. In section 4, we prove that the adversarial attack is more beneficial than pure random walk in locating high-residual regions and give an upper bound for the generalization error of the model under WbAR. In section 5, we conduct several numerical experiments to demonstrate the effectiveness WbAR. Finally, in section 6, we provide conclusions and discussions on further works. 

\section{Preliminaries}
In this section, we provide a brief review of adversarial attacks, PINNs, and some current typical residual-based adaptive sampling strategies. 
\subsection{Adversarial attacks}

Locating samples that fail the current model approximation is a well-studied topic in deep learning known as adversarial attacks. 
Adversarial attacks are tiny perturbations added to the original sample that can lead to neural network misclassification or regression failure \cite{ning2023evaluating}. 
While deep neural networks have demonstrated impressive generalization across different tasks, they are frequently vulnerable to adversarial samples \cite{szegedy2013intriguing}. 

Adversarial attacks can be divided into white-box attacks \cite{szegedy2013intriguing}\cite{goodfellow2014explaining}\cite{kurakin2016adversarial}\cite{carlini2017towards}\cite{moosavi2016deepfool}, for which the attacker knows the parameters and gradients of the model, and black-box attacks \cite{papernot2017practical}\cite{ilyas2018black}\cite{wierstra2014natural}\cite{su2019one}, for which the attacker can only access the input and output of the model. 
In terms of the attacker's objective, adversarial attacks can be classified into targeted attacks and non-targeted attacks.
Targeted attacks means that the attacker tricks the model to misclassify the original sample into a specified class. 
On the other hand, non-targeted attacks only require the model to fail the task. 
For a regression problem that approximates a function $r(x)\in\mathbb{R}$ with $r(x;\theta)\in\mathbb{R}$, targeted adversarial attack is an optimization problem that
\begin{align}\label{eq:opt_attack}
\max_\zeta&\;\;{\mathcal{L} \big(r(x+\zeta;\theta), r(x+\zeta)\big)} \notag \\
s.t.& \;\;\| \zeta \|_\infty \leq \epsilon 
\end{align}
where $\mathcal{L}(\cdot)$ is a distance function of $r(x+\zeta;\theta)$ and $r(x+\zeta)$, and $\epsilon$ is the threshold of maximum allowable adversarial perturbation. 

If the structure and specific parameters of each layer of the model are known, we can apply white-box attacks to generate adversarial samples with a very high success rate. 
One efficient way of solving (\ref{eq:opt_attack}) is through gradient-based methods. 
Madry et. al. \cite{madry2017towards} proposed an iterative gradient-based method named the projected gradient descent (PGD) attack. 
PGD is a numerical approach of solving constrained optimization problems. 
For $x\in\mathbb{R}^n$, $\emptyset \neq \mathcal{Q} \subset \mathbb{R}^n$, PGD solves
\begin{align}
\min_{x\in\mathcal{Q}} f(x)\notag
\end{align}
by iterating the following equation until a stopping condition is met:
\begin{align}
x_{k+1} = \mathcal{P}_{\mathcal{Q}}\left( x_k - \eta_k \nabla f(x_k) \right)\notag
\end{align}
where $k\in\mathbb{N}$ is the current iteration, $\nabla f$ is the gradient of $f$ w.r.t. differentiation of $x$, $\eta_k\in(0,\infty)$ is the gradient stepsize, and $\mathcal{P}_{\mathcal{Q}}(\cdot):\mathbb{R}^n\rightrightarrows\mathbb{R}^n$ is a projection from $\mathbb{R}^n$ to $\mathcal{Q}$. 
PGD is a simple algorithm that if the point $x_k - \eta_k \nabla f(x_k)$ after the gradient update is leaving the set $\mathcal{Q}$, then project it back; otherwise keep the point. 
In the problem (\ref{eq:opt_attack}), the set $\mathcal{Q}$ is constrained by $L^\infty$ norm, thus the projection
\begin{align}
\mathcal{P}_{\mathcal{Q}}(x) = clip_{[-\epsilon,\epsilon]}(x)\notag
\end{align}
where $clip_{[-\epsilon,\epsilon]}(x)$ is clipping every dimensions of $x$ within $[-\epsilon,\epsilon]$. 
The PGD attack method \cite{madry2017towards} also induced a random initialization that adds a uniformly distributed random variable $\mathcal{U}[-\epsilon, \epsilon]$ to $x$ before the first iteration. 
Random initialization helps to prevent the formation of local minima on the training samples during model training.

Note that although PGD attack is particular mentioned in this paper, non-gradient based white-box attacks, for example L-BFGS\cite{szegedy2013intriguing} and C\&W\cite{madry2017towards}, are also powerful adversarial attack methods.

One of the most helpful strategy to defense the adversarial attack is the adversarial training. Adversarial training is simply adding the adversarial samples to the training dataset to improve the model robustness to the attack. Adding the attack into the consideration, the adversarial training is a min-max process
\begin{align}
\theta^* = \arg\min_\theta \mathbb{E}_{\mathcal{Q}}\big( \max_{\zeta, \;\; \| \zeta \|_\infty \leq \epsilon}&\;\;{\mathcal{L} \big(r(x+\zeta;\theta), r(x+\zeta)\big)} \big) \notag
\end{align}

\subsection{PINNs}
Consider partial differential equations:
\begin{align}\label{eq:pde}
\mathcal{A}(x; u(x))=0, x\in\Omega, \notag \\
\mathcal{B}(x; u(x))=0, x\in\partial\Omega
\end{align}
where $\Omega\in\mathbb{R}^d$ is the spatial domain, $x\in\Omega$ is the spatial variable, $\mathcal{A}$ and $\mathcal{B}$ are linear or non-linear differential operator and boundary operator respectively, and $u(x)$ is the unknown PDE solution. 
In PINNs, the time $t$ can be treated as one of the dimensions in the spatial domain.

\begin{figure}[!ht]
\centering
\includegraphics[width=.45 \textwidth]{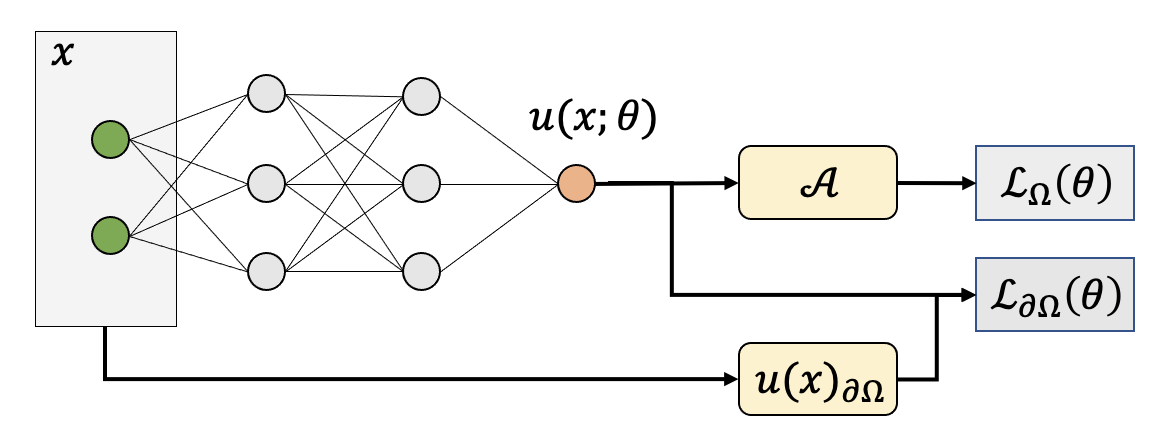}
\caption{The training framework of PINN}
\label{fig:pinn}
\end{figure}

PINN is a deep neural work $u(x;\theta)$ with parameters $\theta$ to approximate $u(x)$. The optimal parameters $\theta^*$ are located by solving the following optimization problem:
\begin{align}\label{eq:nn_opt}
\theta^* &= \arg\min_{\theta\in\Theta} \mathcal{L}(\theta) \notag \\
&= \arg\min_{\theta\in\Theta} \mathcal{L}_\Omega(\theta) + \lambda \mathcal{L}_{\partial\Omega}(\theta) 
\end{align}
where $\Theta$ is the parameter space, $\lambda$ is a trade-off factor balancing the PDE loss $\mathcal{L}_\Omega(\theta)$ and the boundary loss $\mathcal{L}_{\partial\Omega}(\theta)$, and $\mathcal{L}$ is the overall loss of PINN. As shown in Figure \ref{fig:pinn}, the PDE loss $\mathcal{L}_\Omega(\theta)$ usually follows a self-supervised training manner with $L^2$ loss, i.e.,
\begin{align}
\mathcal{L}_\Omega(\theta) = \| \mathcal{A}(x; u(x;\theta)) \|_{2,\Omega}^2 \notag
\end{align}
The operator $\mathcal{A}$ can be implemented through the automatic differentiation engine of deep neural networks. 
The boundary loss $\mathcal{L}_{\partial\Omega}(\theta)$ is commonly derived through supervised training using the $L^2$ loss, i.e.,
\begin{align}
\mathcal{L}_{\partial\Omega}(\theta) = \| \mathcal{B}(x; u(x;\theta)) \|_{2,\partial \Omega}^2 = \| u(x;\theta) - u(x) \|_{2,\partial \Omega}^2 \notag
\end{align}
Here, $u(x)$ is the ground truth solution of the problem on the boundary. Note that, for $x$ on the boundary, it can be included in both $\mathcal{L}_\Omega(\theta)$ and $\mathcal{L}_{\partial\Omega}(\theta)$.  Denote $r(x; \theta)=|\mathcal{A}(x; u(x;\theta))|$ as the approximation residual of the PINN. The above $L^2$-norm is defined as 
\begin{align}
\| r \|^2_{2,\Omega} = \int_\Omega r(x;\theta)^2 \omega(x) dx\notag
\end{align}
where $\omega(x)$ is the prior distribution of the problem. 

In the actual neural network training scheme, the optimization problem (\ref{eq:nn_opt}) is reconstructed following a supervised learning framework.
Instead of being directly calculated on the entire $\Omega$, the loss function is approximated using a training dataset collected according to the prior distribution on $\Omega$. 
The training dataset consists of collocation points $\mathcal{D}=\{s^{(n)}\}_{n=1}^{N}$ and boundary points $\mathcal{D}_b=\{s_b^{(n)}\}_{n=1}^{N_b}$. 
The loss function is the empirical risk calculated on the training dataset, i.e.,
\begin{align}
\hat{\mathcal{L}}(\theta) = \hat{\mathcal{L}}_\Omega(\theta) + \lambda \hat{\mathcal{L}}_{\partial\Omega}(\theta) \notag
\end{align}
where
\begin{align}
&\hat{\mathcal{L}}_\Omega(\theta) = \frac{1}{N} \sum_{n=1}^{N} \| r(s^{(n)};\theta)\|_2^2 \notag\\ 
&\hat{\mathcal{L}}_{\partial\Omega}(\theta) = \frac{1}{N_b} \sum_{n=1}^{N_b} \| u(s_b^{(n)};\theta) - u(s_b^{(n)}) \|_2^2 \notag
\end{align}
Note that, as discussed above, $\mathcal{D}_b$ can be a subset of $\mathcal{D}$. Although $\mathcal{D}$ and $\mathcal{D}_b$ are unbiased samples collected from $\omega(x)$, there is still a gap between the generalization risk $\mathcal{L}(\theta)$ and the empirical risk $\hat{\mathcal{L}}(\theta)$. The gap is
\begin{align}
\mathcal{G}(\theta) &= |\mathcal{L}(\theta) - \hat{\mathcal{L}}(\theta)| \notag\\
&= |\mathcal{L}_\Omega(\theta) + \lambda\mathcal{L}_{\partial\Omega}(\theta) - (\hat{\mathcal{L}}_\Omega(\theta) + \lambda\hat{\mathcal{L}}_{\partial\Omega}(\theta))| \notag \\
&\leq |\mathcal{L}_\Omega(\theta) - \hat{\mathcal{L}}_\Omega(\theta)|+ \lambda|\mathcal{L}_{\partial\Omega}(\theta)  - \hat{\mathcal{L}}_{\partial\Omega}(\theta)|\notag
\end{align}
Consider the gap between the generalization risk and the empirical risk of the PDE loss for now. 
Since $\{s^{(n)}\}_{n=1}^{N}$ are i.i.d. samples drawn from $\omega(x)$, and assume $a\leq r(s^{(n)};\theta)\leq b$, $n=1,\hdots,N$, for a small positive non-zero value $\epsilon$, by Heoffding’s inequality, there holds
\begin{align}
\mathbb{P}\left[ \left| \mathcal{L}_\Omega(\theta) - \hat{\mathcal{L}}_\Omega(\theta) \right|>\epsilon \right] \leq 2 \exp \left(\frac{-2N\epsilon^2}{(b-a)^2}\right)\notag
\end{align}
If the i.i.d. drawn sample has a smaller residual range $b-a$, the upper bound will be tighter. Thus, locating high-residual samples and refining high-residual regions by training can be beneficial in shrinking the gap.

\subsection{Collocation points sampling strategies and iterative training }

As mentioned in section \ref{sec:related_works}, the adversarial training can be related to collocation points sampling strategies. 
In this section, we will discuss different characteristics of different sampling strategies.

Samples can be drawn independently from the model approximation status, for example, the \textit{uniform sampling} where samples are independently and identical distributed (i.i.d.) with $s^{(n)} \sim \mathcal{U}[\Omega]$. 
Suppose we have a model that is already trained on a dataset but fails only in a tiny region. 
If we try to refine the model with samples drawn by uniform sampling, there will be a low probability that samples will fall into the failure region. 
However, new samples in the failure region can help further improve the model while samples in well-approximated regions have much less significance as the model is already well-trained there.
Thus, accurately locating failure regions, especially locating crucial collocation points, can improve the approximation accuracy and the training efficiency. 
In the PINNs framework, the failure region is located through the residual.

Generating more samples in high-residual regions can be achieved through two main approaches: residual distribution approximation and sample refinement. 

\textit{Residual distribution approximation:} 
Although the model residual is accessible on the entire problem domain, locating high-residual regions is challenging due to the non-convex nature of the neural network.
One effective way is to approximate the residual map with an artificial distribution then draw samples from the approximated distribution.
This ensures that samples have a higher probability of being generated in high-residual regions. 
For example, Gao et. al. \cite{gao2023failure} propose Gaussian or Gaussian mixture models to approximate the residual map, which is called \textit{Self-adaptive importance sampling (SAIS)}. 
SAIS begins with a random initialization and iteratively adds new samples which follow Gaussian distributions $\mathcal{N}(\mu, \Sigma)$. 
The Gaussian distributions are estimated from the sampling distribution of the residual and then truncated to be restricted to the problem domain. 
\begin{align}
\mu &= \frac{1}{N_p} \sum_{n=1}^{N_p} \tilde{s}^{(n)} \notag \\
\Sigma &= \frac{1}{N_p-1} \sum_{n=1}^{N_p}(\tilde{s}^{(n)}- \mu) \otimes (\tilde{s}^{(n)} - \mu) \notag
\end{align}
where $\tilde{s}^{(n)}$ are samples with top residuals generated from the Gaussian distribution estimated in the previous iteration. 
Correct residual distribution approximation can increase the probability of the samples falling into high-residual regions. 
However, selecting a suitable distribution in prior can be challenging. 
SAIS assumes that the distribution follows a Gaussian distribution or a mixture of Gaussians, making it suitable for high-residual regions that are shaped like ellipses or collections of ellipses. 

One notable extension of FI-PINN, incorporating an annealing re-sampling technique and subset simulation for failure probability approximation, was subsequently developed in \cite{GTYZ23} and termed \textit{FIPINN-R}. This enhanced approach maintains a fixed number of collocation points throughout training, strategically composed of both uniformly sampled points and samples drawn from an estimated failure distribution. As training progresses, FIPINN-R dynamically adjusts the sampling strategy by progressively reducing the proportion of points drawn from the failure distribution. The method employs subset simulation with Metropolis-Hastings sampling to estimate the failure distribution, offering a more accurate representation compared to conventional Gaussian distribution assumptions.

\textit{Sample refinement:} 
An alternative approach to placing samples in high-residual regions is to directly move samples towards regions of high-residual. 
One of the representative methods is RAR \cite{LMMK21}. 
RAR selects samples with the top-$N$ residuals from a larger sample set generated by uniform sampling, i.e.,
\begin{align}
\{s^{(n)}\}_{n=1}^N = \{\arg\max_x r(x;\theta)\}_{N}, \forall x \in \{\tilde{s}^{(m)}\}_{m=1}^{\lfloor KN \rfloor}\notag
\end{align}
where $K>1$ and $\tilde{s}^{(m)}\sim \mathcal{U}[\Omega]$. 
Sample refinement should find high-residual regions or directions towards high-residual regions accurately and efficiently. 
However this search is often effected by the volume of the region and the dimensions of the problem domain. 
RAR selects high-residual samples from i.i.d. samples, resulting in fewer samples falling into the high-residual region if the volume of the region decreases. 
By the large deviations theory, the number of candidate i.i.d. samples has to exponentially increase to maintain the same probability.

The above sampling strategies can be iteratively applied during the training process. Particularly, samples are iteratively generated based on the current approximation status of the model and appended to the training dataset for additional rounds of training \cite{gao2023failure}\cite{LMMK21}.
In this case, the collocation points are partitioned as  $\mathcal{D}=\cup_{k=1}^K \mathcal{D}_{k}$, where $\mathcal{D}_{k}=\{s_{k}^{(n)}\}_{n=1}^{N_{k}}$ are samples generated in the $k$-th training iteration. 
The model parameters obtained in the previous iteration are loaded and further retrained on $\cup_{i=1}^k \mathcal{D}_{i}$ in the $k$-th training iteration. 
The empirical risk of the collocation points is
\begin{align}
\hat{\mathcal{L}}(\theta) = \frac{\sum_{i=1}^k \sum_{n=1}^{N_i} \| r(s_i^{(n)};\theta)\|_2^2}{\sum_{i=1}^k N_i}  \notag
\end{align}

\section{White-box Adversarial Attack Refinement for PINNs}
\subsection{Motivation}\label{sec:motivation}
Adversarial attacks can move samples towards failure regions by solving the optimization problem (\ref{eq:opt_attack}). 
For PINNs, the equation approximation is preceded through maximizing the residual $r(x;\theta)$. 
Thus, an adversarial attack can be constructed by moving samples towards higher residuals, particularly towards the local maximums of the residuals with sufficient steps. 
Since adversarial samples move towards higher residuals, they can automatically and accurately adapt to the failure regions. 
The adaptation of failure location is independent of the volume of the failure region, the dimension of the problem, or the distribution of the residual map, which is especially helpful in solving PDEs with specific characteristics such as sharp, oscillatory or multi-peak solutions, etc.

After generating adversarial samples, we apply adversarial training to force the model to focus on the failure regions, i.e. refine the model with the adversarial samples. 
Further, under the iterative training framework, adversarial attacks can be applied in each training iteration to generate samples surrounding the local maximums of the residual. 
Then, the residual peaks will be suppressed during the adversarial training and new peaks will be captured in the next training iteration. 
The adversarial attack helps to carefully select samples around, if not exactly at, the local maximums, thus the residual should be better suppressed than taking samples at random positions. 
Moreover, the procedure is adaptive to all kinds of residual maps, which can provide better suitability for various complicated equations.

In the following sections, we will describe the adversarial attack algorithm, named PINN-PGD, in detail. 
The PINN-PGD is particularly designed to generate adversarial samples for PINNs based on an $L^\infty$ constrained PGD attack method. 
Then, we illustrate the iterative white-box adversarial attack refinement framework, named WbAR. 
The WbAR extends the basic adversarial training with sample reinforcement augmentation and revisiting as hyperparameters to control the sampling during the training iteration. 
The effects of all the hyperparameters will be discussed and verified in section \ref{sec:hyperparam}.

\subsection{PINN-PGD}

For the regression problem of (\ref{eq:opt_attack}), we extends the basic PGD attack as shown in Algorithm \ref{alg:r-pgd}. 
In the random initialization step, a uniformly distributed random variable $\mathcal{U}[-\epsilon, \epsilon]$ is added to the original sample $x$. 
Then, the perturbed sample is clipped back to the problem domain $\Omega$.
In the projected gradient descent process, as we are maximizing the residual, gradient ascent is applied. 
The model takes a step of $\eta$ in the direction of the loss gradient each time. 
  The overall perturbation is clipped to stay within the maximum perturbation threshold $[-\epsilon, \epsilon]$.
Finally, the perturbed sample is clipped to ensure that the adversarial sample still belongs to $\Omega$. 

\begin{algorithm}
\caption{PINN-PGD}\label{alg:r-pgd}
\begin{algorithmic}[1]
\REQUIRE {The original sample $x$, the problem domain $\Omega$, the regression model $r(x;\theta)$, the target function $r(x)$, the loss $\mathcal{L}(\cdot)$, the maximum perturbation threshold $\epsilon$, number of iteration steps $T$, and the step size of each iteration $\eta$.}
\STATE {$g_0 \gets 0$}
\STATE {$x \gets x + \mathcal{U}[-\epsilon, \epsilon]$ \hfill$\triangleright${ Random initialization}}
\STATE {$x \gets clip_\Omega(x)$}
\FOR{$t = 1:T$}
	\STATE {$g_x \gets \eta * sign\big(\nabla_x r(x;\theta)\big)$}\hfill$\triangleright${ Gradient ascent}
	\STATE {$g_0 \gets clip_{[-\epsilon, \epsilon]}(g_0 + g_x)$}
	\STATE {$x \gets x + g_0$}
\ENDFOR
\STATE {$x \gets clip_\Omega(x)$}
\end{algorithmic}
\end{algorithm}

\begin{figure}[!t]
\centering
\includegraphics[width=.45\textwidth]{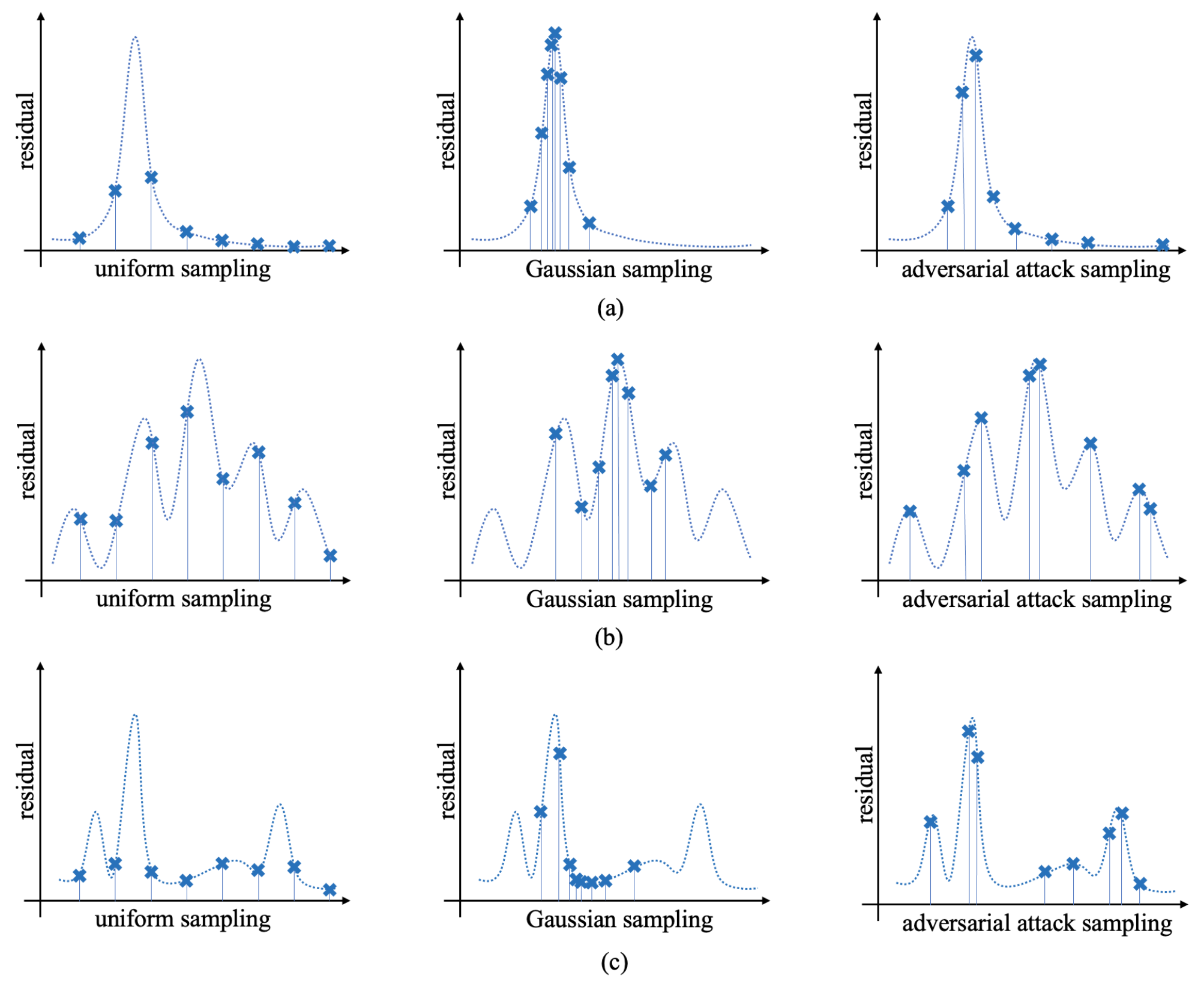}
\caption{The residual and samples of (a) a sharp solution, (b) an oscillatory solution, (c) a multi-peak solution problem.}
\label{fig:motivation}
\end{figure}

\subsection{WbAR}

WbAR follows the iterative training framework. 
At initialization, $N_0$ samples $\{s_0^{(n)}\}_{n=1}^{N_0}$ are generated through the LHS. Then, at the iteration $k$, we calculate the adversarial samples $\{a^{(n)}_k\}_{n=1}^{N_{k-1}}$,$\hdots$, $\{a^{(n)}_{k-m}\}_{n=1}^{N_{k-m-1}}, \{a_1^{(n)}\}_{n=1}^{N_0}$  for all the samples generated in the previous $m$ iterations and the initial samples, i.e., $\{s^{(n)}_{k-1}\}_{n=1}^{N_{k-1}}$, $\hdots$, $\{s^{(n)}_{k-m-1}\}_{n=1}^{N_{k-m-1}}, \{s_0^{(n)}\}_{n=1}^{N_0}$. 
The calculated adversarial samples are used as candidate samples. 
New samples are selected from the $N_k$ highest residual candidate samples, where $N_k$ is the number of samples of the $k$-th iteration. 
The purpose of revisiting the samples in the previous $m$ iterations is to speed up the convergence of the samples to the high residual areas. 
And revisiting the initialization samples helps avoid the iteration getting stuck in local regions. The model is then retrained on new samples before the next adversarial training iteration starts. 
The details can be found in Algorithm \ref{alg:aaas}.

From an optimization perspective, WbAR is similar as local maximum searching for non-convex function using a gradient-based method from multiple initial points. 
The difference is that WbAR has constrains on the number of iteration steps and the search  distance, which maintains the samples surrounding the maximums instead of converging at them.

\subsection{Sampling Strategies Comparison}

As illustrated in Figure \ref{fig:motivation}, if the model has a sharp solution, it usually has one dominant peak in the residual. 
Since the area of the residual peak can be very small, the probability of samples falling onto the peak will be low if a uniform sampling strategy is taken. 
If an adaptive Gaussian sampling is applied, the samples will gather around the peak and leave the remaining areas of the problem domain less sampled. 
If we use uniform sampling as initialization points (as shown in the first column of Figure \ref{fig:motivation}) and apply an adversarial attack to generate samples, the samples near the peak will move to the peak while samples in the remaining areas will not move significantly. 
For the problems with oscillatory solutions, uniform sampling may fail to capture every peak accurately, while adaptive Gaussian sampling might miss peaks that are distant from the mean of the distribution.
However, adversarial attack sampling can capture every peaks as long as samples are initialized near each peak. 
As demonstrated in Figure \ref{fig:motivation} (b), when we took uniform sampling as the initialization, adversarial attack can locate all the peaks in the residual map. 
The situation is similar in the multi-peak solution problem, as shown in Figure \ref{fig:motivation} (c). Adversarial samples can effectively capture the distribution of the residual, even in cases where the distribution is complicated to be expressed explicitly.

\begin{algorithm}
\small
\caption{WbAR}\label{alg:aaas}
\begin{algorithmic}[1]
\REQUIRE The model $u(x;\theta)$, the residual $r(x;\theta)$ of model, the number of training iterations $K$, the number of sampling points in each iteration $N_k$, $k=0,\hdots,K$, the problem domain $\Omega$, the maximum perturbation threshold $\epsilon$, the number of iteration steps $T$, the step size of each iteration $\eta$, and the number of revisiting iterations $m$.
\STATE Sample $\{s_0^{(n)}\}_{n=1}^{N_0}$ through LHS
\STATE $u(x;\theta)\gets$ train on $\{s_0^{(n)}\}_{n=1}^{N_0}$
\FOR{$k=1:K$}
	\FOR{$i=(k-1-\floor{m}):(k-1)$}
	\STATE $\{a_{i+1}^{(n)}\}_{n=1}^{N_i} \gets \{\text{PINN-PGD}(s_{i}^{(n)},|r(x;\theta)|, \epsilon, \eta, T, \Omega)\}_{n=1}^{N_i}$
	\ENDFOR
	\IF{$m>\floor{m}$}
	\STATE $\{a_{k-\ceil{m}}^{(n)}\}_{n=1}^{\floor{(m-\floor{m})*N_{k-\floor{m}}}} \gets$\par$ \{\text{PINN-PGD}(s_{k-\floor{m}}^{(n)},|r(x;\theta)|, \epsilon, \eta, T, \Omega)\}_{n=1}^{\floor{(m-\floor{m})*N_{k-\floor{m}}}}$
	\ENDIF
	\STATE $\{a_{1}^{(n)}\}_{n=1}^{N_i} \gets \{\text{PINN-PGD}(s_{0}^{(n)},|r(x;\theta)|, \epsilon, \eta, T, \Omega)\}_{n=1}^{N_0}$
	\STATE $\{s_{k}^{(n)}\}_{n=1}^{N_k} \gets$ $\{\arg\max_x r(x;\theta)\}_{N_k}$, \par$\forall x \in \left(\cup_{i=k-\ceil{m}}^{k}\{a_{i}^{(n)}\}\right)\cup\{a_1^{(n)}\} $ 
	\STATE $u(x;\theta)\gets$ retrain on $\left(\cup_{i=0}^k\{s_{i}^{(n)}\}\right)$
\ENDFOR
\end{algorithmic}
\end{algorithm}

\section{Convergence Analysis of WbAR}
In this section, we first prove the adversarial attack is more beneficial than pure random walk in locating high-residual regions (Theorem \ref{thm:better}), and then give an upper bound for the generalization error of the model under WbAR (Theorem \ref{thm:bound}). The following assumptions are necessary:
\begin{assumption}\label{ass:r}
For the residual $r(x;\theta)$ of PINN, assume
\begin{enumerate}[label=(\roman*)]
\item $r(x;\theta)$ is continuous on $\Omega$;
\item $r(s_k^{(n)};\theta) \geq r(\tilde{s}_k^{(n)};\theta)$, $\forall k=1,\hdots, K$, $\forall n=1,\hdots, N_k$;
\item $\exists \delta>0$, s.t. $r(x;\theta) \geq r(\tilde{x};\theta)$, $\forall x \in C_\delta(s_k^{(n)})$, $\forall \tilde{x} \in C_\delta(\tilde{s}_k^{(n)})$,
\end{enumerate}
where $s_k^{(n)}$ is the $n$-th adversarial sample generated in the $k$-th iteration, $\tilde{s}_k^{(n)}$ is the corresponding sample position before the adversarial attack gradient ascent step, $C_\delta(s_k^{(n)})$ and $C_\delta(\tilde{s}_k^{(n)})$ are hypercube neighborhoods of side length $\delta$ centered at $s_k^{(n)}$ and $\tilde{s}_k^{(n)}$, respectively.
\end{assumption}

Assumption \ref{ass:r}(i) gives a mild assumption of the continuity of the model residual. 
For Assumption \ref{ass:r}(ii), as the projected gradient ascent steps are applied to $r(x;\theta)$ at  $x = \tilde{s}_k^{(n)}$, we assume $r(s_k^{(n)};\theta) \geq r(\tilde{s}_k^{(n)};\theta)$. Assumption \ref{ass:r}(iii) automatically holds by (i) and (ii). 
The neighborhoods are described as hypercubes, because the $L_\infty$ norm is applied as constrains in the adversarial attack.

\begin{theorem}\label{thm:better}
Suppose Assumption \ref{ass:r}(ii) is satisfied. 
Let $r_{k-1}(x;\theta^*)$ be the model residual obtained after $k-1$ training iterations, the empirical risk of samples $\{s^{(n)}_k\}_{n=1}^{N_k}$ will be greater or equal to the empirical risk of samples $\{\tilde{s}^{(n)}_k\}_{n=1}^{N_k}$, i.e.,
\begin{align}
\sum_{n=1}^{N_k} r_{k-1}(s^{(n)}_k;\theta^*)^2 \geq \sum_{n=1}^{N_k} r_{k-1}(\tilde{s}^{(n)}_k;\theta^*)^2
\end{align}
\end{theorem}
\begin{proof}
By Assumption \ref{ass:r}(ii), we have 
$$r_{k-1}(s^{(n)}_k;\theta^*)\geq r_{k-1}(\tilde{s}^{(n)}_k;\theta^*)$$
Combining the independent $N_k$ samples, the theorem can be proved.
\end{proof}
Theorem \ref{thm:better} claims that, before each new training iteration, the gradient ascent step can locate the samples on regions with higher residuals. 
The higher-residual regions will be further suppressed in the next training iteration. 
Thus, the gradient ascent can help locating high-residual regions.

\begin{assumption}\label{ass:ineq} \cite{gao2023failure} 
Let $\mathcal{A}$ be a linear operator that maps $\mathcal{X} \rightarrow \mathcal{X}$ in a partial differential equation as (\ref{eq:pde}), where $\mathcal{X}$ is a Hilbert space. We assume that the operator $\mathcal{A}$ and the boundary operator $\mathcal{B}$ satisfy
\begin{equation}
\alpha \|v\|_{2,\Omega} \leq \|\mathcal{A}v \|_{2,\Omega} + \|\mathcal{B}\|_{2,\partial\Omega}, \; \forall v\in \mathcal{X} \notag
\end{equation}
where $\alpha$ is positive constant which is independent of $v$.
\end{assumption}

\begin{assumption} \label{ass:bound}
Suppose the PINN model $u(x; \theta)$ is over-parameterized and sufficiently trained on training samples $\{s^{(n)}_k\}_{n=1}^{N_k}$, 
\begin{enumerate}[label=(\roman*)]
\item $\exists \tau>0$, s.t. $\tau := \sup_{x\in\Gamma} |r(x; \theta^*)| <\infty$;
\item $\exists \tau_b>0$, s.t. $\tau_b:= \max_{x\in\partial\Omega} \|\mathcal{B}(u(x) - u(x;\theta^*)) \|_{2,\partial\Omega}<\infty$;
\item $\exists M>0$, s.t. $M := \sup_{x\in\Omega} |r(x; \theta^*)| <\infty$,
\end{enumerate}
\end{assumption}

The PINN model is trained by minimizing the mean-squared residual (PDE loss). 
If the model is over-parameterized and sufficiently trained, residuals on every training samples will converge to zero. 
By the continuity of the residual, the Assumption \ref{ass:bound}(i) holds. 
Similarly, (ii) holds by the boundary loss and (iii) holds by the continuity.

\begin{assumption}\label{ass:attack}
The displacement during gradient ascent of an adversarial attack is negligible comparied to the displacement that occurs during random initialization.
\end{assumption}

The displacement during gradient ascent only depends on the initialization position and the model residual. 
In other words, it is conditionally independent from all the previous adversarial attack iterations given the its initialization position. 
For simpler analysis, the displacement during gradient ascent can be ignored without effecting the convergence analysis result. 
Thus, the adversarial sample of $s_k^{(n)}$ can be approximated at $\tilde{s}_k^{(n)}$ for Theorem \ref{thm:bound}.

\begin{lemma}\label{lemma:init}
Denoting the hypercube neighborhood of the samples $\tilde{s}_k^{(n)}$ as $C_{\delta}(\tilde{s}_k^{(n)})$, for any point $x$ in $\Omega$, the probability that it falls in the hypercubes of two adjacent adversarial iterations is
\begin{align}
P(x\in \cup_{i=k-1}^{k} C_{\delta}(\tilde{s}_i^{(n)})) = 2p - q_1
\end{align}
where, $p={V(C_\delta)}\slash{V(\Omega)}\ll 1$,  $q_1={V(C_\delta)}\slash{V(C_{2\epsilon})}*p\ll 1$, $V(C_l)$ is the volume of the hypercube of side length $l$, and $\epsilon$ is the maximum perturbation of the adversarial attack.
\end{lemma}

\begin{proof}
For simpler notation, we denote $C_i^{(n)} := C_{\delta}(\tilde{s}_i^{(n)})$.
By law of total probability,
\begin{align}\label{eq:lemma1_total_prob}
&P(x\in \cup_{i=k-1}^{k} C_i^{(n)}) \notag \\
=& P(x\in \cup_{i=k-1}^{k} C_i^{(n)} | \cap_{i=k-1}^{k} C_i^{(n)}=\emptyset)* \notag \\
&P( \cap_{i=k-1}^{k} C_i^{(n)}=\emptyset) + \notag \\
&P(x\in \cup_{i=k-1}^{k} C_i^{(n)} | \cap_{i=k-1}^{k} C_i^{(n)}\neq\emptyset)*\notag \\
&P( \cap_{i=k-1}^{k} C_i^{(n)}\neq\emptyset)
\end{align}
With the no intersection condition, 
\begin{align}\label{eq:lemma1_term1}
P(x\in \cup_{i=k-1}^{k} C_i^{(n)} | \cap_{i=k-1}^{k} C_i^{(n)}=\emptyset)=2p
\end{align}
By Assumption \ref{ass:attack}, $\tilde{s}_{k}^{(n)}$ uniformly falls in the hypercube of volume $V(C_{2\epsilon})$ centered on $\tilde{s}_{k-1}^{(n)}$. The probability that two hypercubes of volume $V(C_{\delta})$ have no intersection equals to
\begin{align}\label{eq:lemma1_term2}
P( \cap_{i=k-1}^{k} C_i^{(n)}=\emptyset) = 1-q_2
\end{align}
where $q_2 = {V(C_{2\delta})}\slash{V(C_{2\epsilon})} <1$.

The probability that $x$ falls in intersected hypercubes is
\begin{align}\label{eq:lemma1_term3}
&P(x\in \cup_{i=k-1}^{k} C_i^{(n)} | \cap_{i=k-1}^{k} C_i^{(n)}\neq\emptyset) \notag\\
=&\int_{-\delta}^\delta \cdots \int_{-\delta}^\delta \left(2p-\frac{\prod_{j=1}^d (\delta-|x_j|)}{V(\Omega)}\right)\frac{1}{V(C_{2\delta})} dx_{1\cdots d} \notag\\
=& 2p - \frac{V(C_{\delta/2})}{V(\Omega)}
\end{align}

Substituting Eq. (\ref{eq:lemma1_term1}), Eq. (\ref{eq:lemma1_term2}), and Eq. (\ref{eq:lemma1_term3}) into Eq. (\ref{eq:lemma1_total_prob}), it can be obtained that
\begin{align}
&P(x\in \cup_{i=k-1}^{k} C_i^{(n)}) \notag \\
= &2p*(1- q_2) + (2p - \frac{V(C_{\delta/2})}{V(\Omega)})*q_2 =  2p - q_1
\end{align}
\end{proof}

\begin{lemma}\label{lemma:all}
For any point $x$ in $\Omega$, the probability that it falls in hypercubes of $k+1$ adjacent adversarial iterations is
\begin{align}
P(x\in \cup_{i=0}^{k} C_i^{(n)}) \geq (k-1)(1-q_2-\eta)p + (2p-q_1)
\end{align}
where $q_2={V(C_{2\delta})}\slash{V(C_{2\epsilon})}<1$ and $\eta<1$ is a constant.
\end{lemma}

\begin{proof}
Similarly, by law of total probability,
\begin{align}
&P(x\in \cup_{i=0}^{k} C_i^{(n)})  \notag\\
= &P\left(x\in \cup_{i=0}^{k} C_i^{(n)} | C_k^{(n)}\cup (\cap_{i=0}^{k-1} C_i^{(n)})=\emptyset\right) \notag \\
&P\left(C_k^{(n)}\cup (\cap_{i=0}^{k-1} C_i^{(n)})=\emptyset\right)+ \notag \\
&\;\;\;\;\;P\left(x\in \cup_{i=0}^{k} C_i^{(n)} | C_k^{(n)}\cup (\cap_{i=0}^{k-1} C_i^{(n)})\neq\emptyset\right) \notag \\
&P\left(C_k^{(n)}\cup (\cap_{i=0}^{k-1} C_i^{(n)})\neq\emptyset\right) \notag
\end{align}
With the condition that $C_k^{(n)}$ has no intersection with $\cap_{i=0}^{k-1} C_i^{(n)}$, 
\begin{align}
&P\left(x\in \cup_{i=0}^{k} C_i^{(n)} | C_k^{(n)}\cup (\cap_{i=0}^{k-1} C_i^{(n)})=\emptyset\right) \notag \\
= &\left(p+P(x\in \cup_{i=0}^{k-1} C_i^{(n)})\right) \notag
\end{align}
Since the probability of $x$ falling in $k-1$ hypercubes is smaller than the probability of $x$ falling in $k$ hypercubes,
\begin{align}
&P\left(x\in \cup_{i=0}^{k} C_i^{(n)} | C_k^{(n)}\cup (\cap_{i=0}^{k-1} C_i^{(n)})\neq\emptyset\right) \notag \\
&\geq P\left(x\in \cup_{i=0}^{k-1} C_i^{(n)} | C_k^{(n)}\cup (\cap_{i=0}^{k-1} C_i^{(n)})\neq\emptyset\right)\notag
\end{align}
Noticing that $x\in \cup_{i=0}^{k-1} C_i^{(n)}$ no longer depends on $C_k^{(n)}\cup (\cap_{i=0}^{k-1} C_i^{(n)})\neq\emptyset$,  we have the right hand side
\begin{align}
&P\left(x\in \cup_{i=0}^{k-1} C_i^{(n)} | C_k^{(n)}\cup (\cap_{i=0}^{k-1} C_i^{(n)})\neq\emptyset\right) \notag \\
=& P\left(x\in \cup_{i=0}^{k-1} C_i^{(n)}\right)\notag
\end{align}
Thus, 
\begin{align}
&P(x\in \cup_{i=0}^{k} C_i^{(n)}) \notag \\
\geq& \left(p+P(x\in \cup_{i=0}^{k-1} C_i^{(n)})\right)* \notag \\
&P\left(C_k^{(n)}\cup (\cap_{i=0}^{k-1} C_i^{(n)})=\emptyset\right)+ \notag \\
&P\left(x\in \cup_{i=0}^{k-1} C_i^{(n)}\right)*
P\left(C_k^{(n)}\cup (\cap_{i=0}^{k-1} C_i^{(n)})\neq\emptyset\right)\notag
\end{align}
Reordering the inequality, it can be obtained that
\begin{align}\label{eq:lemma2_1}
&P(x\in \cup_{i=0}^{k} C_i^{(n)}) \notag \\
\geq& p*P\left(C_k^{(n)}\cup (\cap_{i=0}^{k-1} C_i^{(n)})=\emptyset\right)+ \notag \\
&P\left(x\in \cup_{i=0}^{k-1} C_i^{(n)} \right) 
\end{align}
By the distributive law of sets,
\begin{align}
&P\left(C_k^{(n)}\cup (\cap_{i=0}^{k-1} C_i^{(n)})=\emptyset\right) \notag \\
\geq& \left(1-\sum_{i=0}^{k-1} P(C_k^{(n)} \cap C_i^{(n)}\neq\emptyset) \right)\notag
\end{align}

By Assumption \ref{ass:attack}, $\tilde{s}^{(n)}_i$ is a random walk with uniform distribution on each step, $P(\tilde{s}_k^{(n)} \in C_{k-i}^{(n)})$ follows Irwin–Hall distribution and exponentially decrease as $i$ increase. Thus, there exists a finite bound $\eta$ such that 
\begin{align}
\sum_{i=0}^{k-2} P(C_k^{(n)} \cap C_i^{(n)}\neq\emptyset) \leq \eta \notag
\end{align}
and $\eta+q_2<1$. Thus,
\begin{align}
P\left(C_k^{(n)}\cup (\cap_{i=0}^{k-1} C_i^{(n)})=\emptyset\right) \geq (1-q_2-\eta) \notag
\end{align}
Equation (\ref{eq:lemma2_1}) is converted into an arithmetic progression,
\begin{align}
P(x\in \cup_{i=0}^{k} C_i^{(n)}) \geq p(1-q_2-\eta) + P(x\in \cup_{i=0}^{k-1} C_i^{(n)}) \notag
\end{align}
By Lemma \ref{lemma:init}, $P(x\in \cup_{i=0}^{1} C_i^{(n)})=2p-q_1$, it can be obtained that
\begin{align}
P(x\in \cup_{i=0}^{k} C_i^{(n)}) \geq (k-1)(1-q_2-\eta)p + (2p-q_1)\notag
\end{align}
\end{proof}

Lemma \ref{lemma:all} claims that initiating from one point, the trajectory and its hypercube neighborhoods after $k$ times WbAR iterations cover at least $(k-1)(1-q_2-\eta)p + (2p-q_1)$ portions of $\Omega$. 
As $k\rightarrow\infty$, WbAR covers entire $\Omega$. 
In the next Lemma, we consider the situation of initiating from $N$ independently sampled points.

\begin{lemma}\label{lemma:n_points}
For any point $x$ in $\Omega$, the probability that it falls in hypercubes, which initiating from $N$ independent points $\{\tilde{s}_0^{(1)}, \tilde{s}_0^{(2)}, \hdots, \tilde{s}_0^{(N)}\}$ and each traversed $k$ adjacent adversarial iterations, is
\begin{align} \notag
P(x\in \cup_{i=0,n=1}^{k,N} C_i^{(n)}) \geq 1 - \big(1- p_a(k)\big)^N
\end{align}
where $p_a(k) = (k-1)(1-q_2-\eta)p - (2p-q_1)$.
\end{lemma}
The proof of Lemma \ref{lemma:n_points} is trivial by the independency of initiating points. Lemma \ref{lemma:n_points} claims that, with $N$ independent trajectories, the trajectory and neighborhoods of WbAR iterations cover larger areas in $\Omega$. As $N\rightarrow\infty$, $P(x\in \cup_{i=0,n=1}^{k,N} C_i^{(n)})\rightarrow 1$.

\begin{theorem}\label{thm:bound}
Assume the problem domain $\Omega$ is bounded, and let $u(x;\theta^*)$ be the problem solution of (\ref{eq:pde}) obtained from the output of a sufficiently trained PINN on a training dataset sampled by WbAR. Suppose Assumptions \ref{ass:r}, \ref{ass:bound}, \ref{ass:ineq} and \ref{ass:attack}, are satisfied. Then the following error estimation holds
\begin{align}
&\| u(x) - u(x;\theta^*) \|_{2,\Omega} \notag \\
&\leq \sqrt{2}\alpha^{-1}\big(\tau^2 + \tau_b^2+ (1-p_a(k))^NM\big)^\frac{1}{2}
\end{align}
where $p_a(k) = (k-1)(1-q_2-\eta)p - (2p-q_1)<1$, $p={V(\delta)}\slash{V(\Omega)}\ll 1$,  $q_1={V(C_{\delta})}\slash{V(C_{2\epsilon})}*p\ll 1$, $q_2={V(C_{2\delta})}\slash{V(C_{2\epsilon})}<1$, $\eta<1$ is a constant, and $q_2+\eta < 1$.
\end{theorem}

\begin{proof}
Let $v=u(x) - u(x;\theta^*)$. By Assumption \ref{ass:ineq}, it can be shown that
\begin{align}
&\| u(x) - u(x;\theta^*) \|_{2,\Omega} \notag \\
 \leq& \alpha^{-1}\big( \|\mathcal{A}(u(x) - u(x;\theta^*)) \|_{2,\Omega} + \notag \\
 &\|\mathcal{B}(u(x) - u(x;\theta^*)) \|_{2,\partial\Omega}  \big) \notag \\
 =& \alpha^{-1}\big( \| r(x;\theta^*) \|_{2,\Omega} + \|\mathcal{B}(u(x) - u(x;\theta^*)) \|_{2,\partial\Omega}  \big) \notag \\
 \leq& \sqrt{2}\alpha^{-1}\big( \| r(x;\theta^*) \|_{2,\Omega}^2 + \|\mathcal{B}(u(x) - u(x;\theta^*)) \|_{2,\partial\Omega}^2  \big)^{\frac{1}{2}} \notag
\end{align}
By Assumption \ref{ass:bound}(ii), we have
\begin{align}\label{eq:thm1_risk_all}
\| u(x) - u(x;\theta^*) \|_{2,\Omega} \leq \sqrt{2}\alpha^{-1}\left( \| r(x;\theta^*) \|_{2,\Omega}^2 + \tau_b^2  \right)^{\frac{1}{2}}
\end{align}
Divie $\Omega$ into $\Gamma_{\delta}:=\cup_{i=0, n=1}^{k,N} C_{\delta}(s_i^{(n)})$ and $\Omega \backslash \Gamma_{\delta}$, where $\Gamma_{\delta}$ is the union of all hypercube neighborhoods of length $\delta$ for every sample in the training dataset. We have
\begin{align}\label{eq:thm1_risk}
&\| r(x;\theta^*) \|_{2,\Omega}^2 \notag \\
=& \int_\Omega r(x;\theta^*)^2 \frac{1}{V(\Omega)}dx \notag \\
= & \int_{\Gamma_{\delta}} r(x;\theta^*)^2 \frac{1}{V(\Omega)}dx + \int_{\Omega\backslash{\Gamma_{\delta}}} r(x;\theta^*)^2 \frac{1}{V(\Omega)}dx 
\end{align}
By Assumption \ref{ass:bound}, if all the WbAR samples are used in training, in their hypercube of length $\delta$ neighborhoods, one has $|r(x;\theta^*)|\leq \tau$. Combining Lemma \ref{lemma:n_points}, the first term of Equation (\ref{eq:thm1_risk}) is bounded by
\begin{align}\label{eq:thm1_term1}
(1-(1-p_a(k))^N) \tau^2 \leq \int_{\Gamma_\delta} r(x;\theta^*)^2\frac{1}{V(\Omega)}dx \leq  \tau^2
\end{align}
By Assumption \ref{ass:bound}, the second term of Equation (\ref{eq:thm1_risk}) is bounded by
\begin{align}\label{eq:thm1_term2}
\int_{\Omega\backslash{\Gamma_{\delta}}} r(x;\theta^*)^2 \frac{1}{V(\Omega)}dx \leq \big(1-p_a(k)\big)^NM
\end{align}
Combining Eq. (\ref{eq:thm1_term1}) and Eq. (\ref{eq:thm1_term2}), it can be obtained that
\begin{align}\label{eq:thm1_res}
\| r(x;\theta^*) \|_{2,\Omega}^2 \leq \tau^2 + \big(1-p_a(k)\big)^NM
\end{align}
By substituting Equation (\ref{eq:thm1_res}) into Equation (\ref{eq:thm1_risk_all}), we have
\begin{align}
&\| u(x) - u(x;\theta^*) \|_{2,\Omega} \notag \\
&\leq \sqrt{2}\alpha^{-1}\big( \tau^2  + \tau_b^2 + (1-p_a(k))^NM \big)^{\frac{1}{2}}
\end{align}
\end{proof}

In Theorem \ref{thm:bound}, we demonstrate that the approximation error on the problem domain $\Omega$ is bounded.  
The upper bound is controlled by the number of  WbAR iterations $k$ and the number of WbAR initial points $N$. 
As shown in Lemma \ref{lemma:all} and Lemma \ref{lemma:n_points}, $p_a(k)$ monotonically increases as $k$ increases, thus $(1-p_a(k))^N$ monotonically decrease as $k$ or $N$ increases. 
In other word, if sufficient initial points and sufficient iterations are taken, WbAR will ensure that the error of equation approximation is bounded. 
However, the upper bound in Theorem \ref{thm:bound} is valid simply because of properties of the random walk rather than adversarial attacks. 
In Theorem \ref{thm:better}, we have shown that adversarial attacks can benefit to locate the high-residual regions than pure random walk.

\section{Numerical Experiments}
\subsection{Two-dimensional Poisson equation}

In the first numerical experiments, we consider the following two-dimensional Poisson equation:
\begin{align}\label{eq:poisson}
-\Delta u(x,y) = f(x,y) \;\;\;\;\; &\text{in } \Omega \notag \\
u(x,y) = g(x,y) \;\;\;\;\; &\text{on } \partial\Omega
\end{align}
where $\Omega=[-1, 1]^2$ and the solution is
\begin{align}
u(x,y) =& exp\big({-100(x-.8)^2}\big) + exp\big({-100x^2}\big) + \notag \\
&exp\big({-100(x+.8)^2}\big)  - exp\big({-100(y-.8)^2}\big) - \notag \\
&exp\big({-100y^2}\big) - exp\big({-100(y+.8)^2}\big)
\end{align}
The true solution is shown in Figure \ref{fig:poisson_gt} which contains three ridges at $x=\{-0.8, 0, 0.8\}$ and three canyons at $y=\{-0.8, 0, 0.8\}$. 

\afterpage{
\begin{figure*}[H]
    \centering
    \begin{subfigure}[b]{.23\textwidth}
    	\centering
	\includegraphics[width=.9\textwidth]{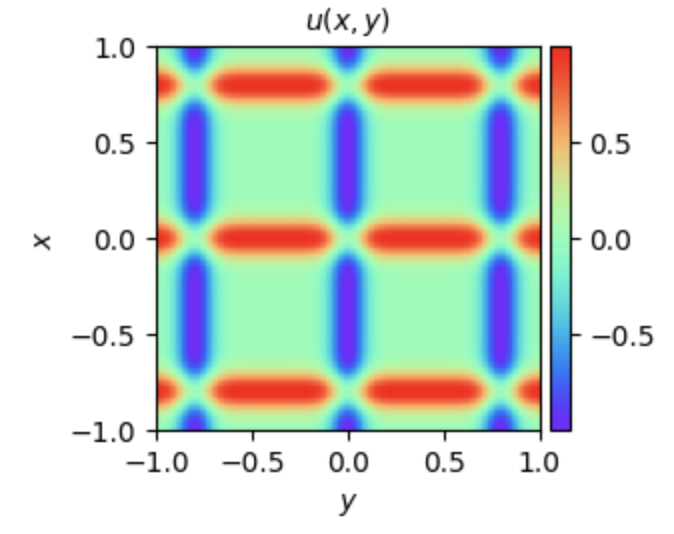}
	\caption{True solution of the two-dimensional Poisson equation}
	\label{fig:poisson_gt}
    \end{subfigure}
    \hfill
    \begin{subfigure}[b]{.73\textwidth}
    	\centering
	\includegraphics[width=.9\textwidth]{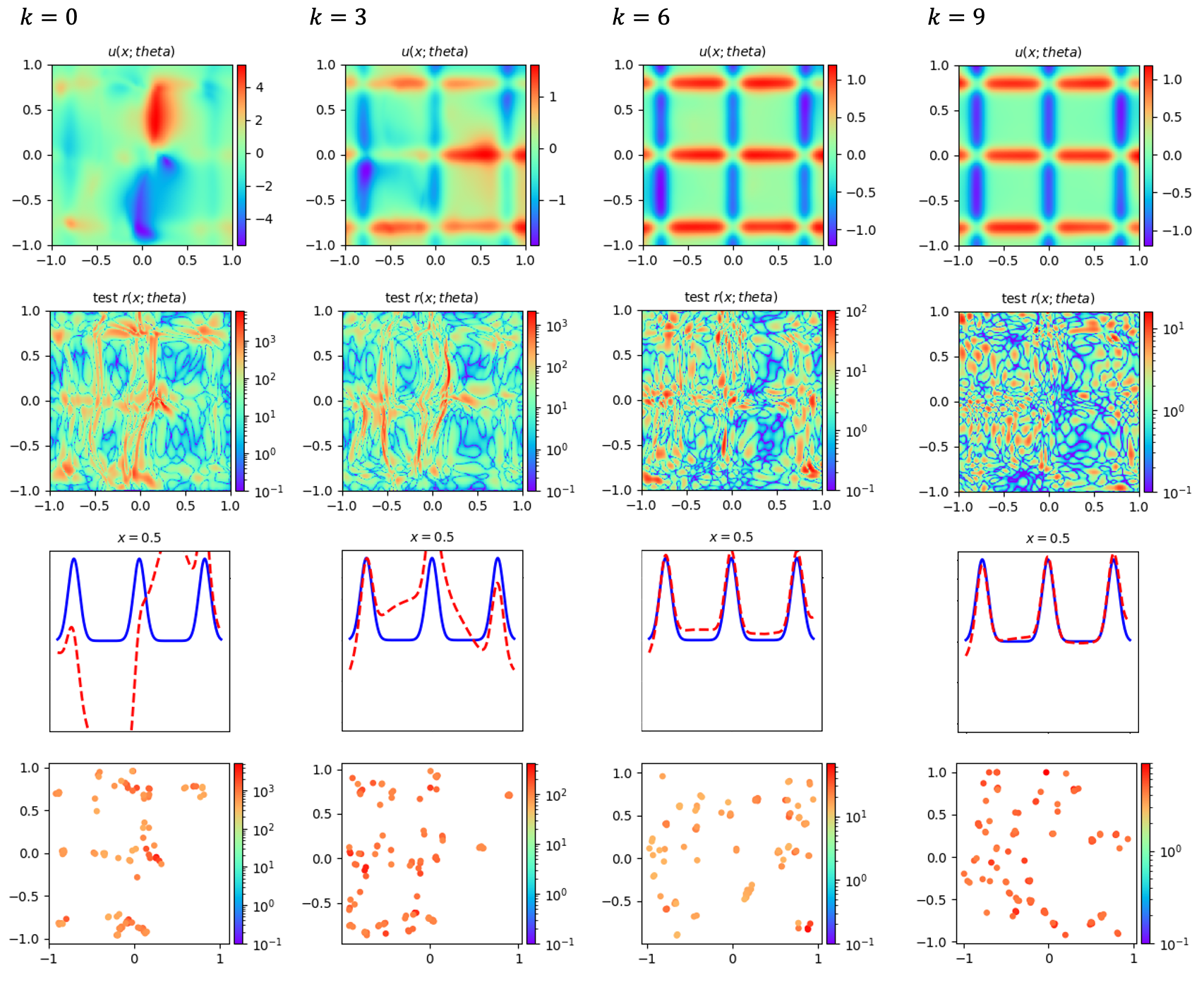}
	\caption{WbAR for the two-dimensional Poisson equation. (1st row) model prediction at the test dataset; (2nd row) model prediction residual;  (3rd row) blue solid: solution at $x=0.5$, red dashed: model prediction; (4th row) samples generated by WbAR.}
	\label{fig:poisson_aaas}
    \end{subfigure}
    \hfill
    \begin{subfigure}[b]{0.48\textwidth}
        \includegraphics[width=\textwidth]{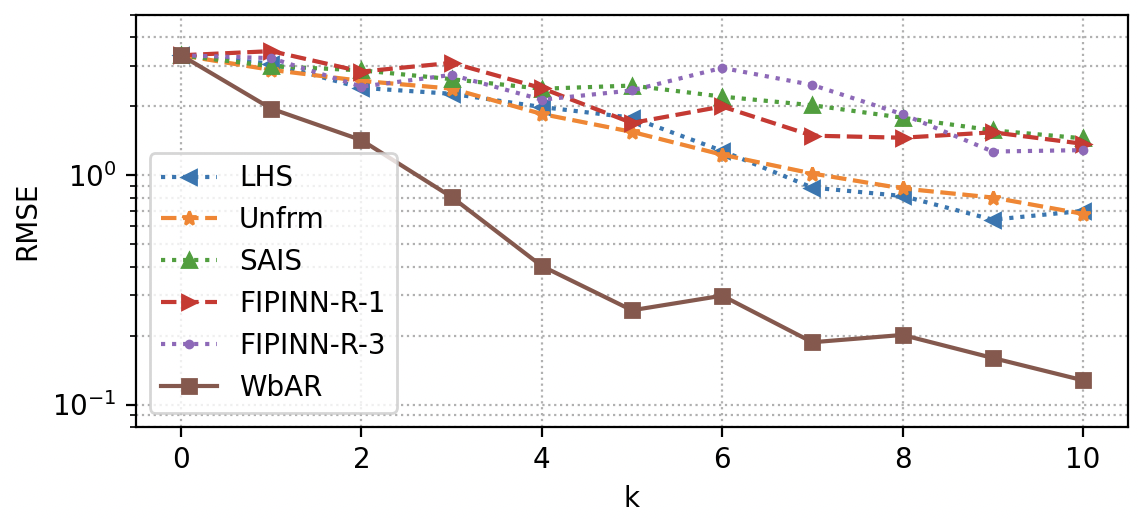}
	\caption{The RMSE of solution on the test dataset for the Poisson equation.}
	\label{fig:poisson_comparision}
    \end{subfigure}
    \hfill
    \begin{subfigure}[b]{0.48\textwidth}
    	\centering
        \includegraphics[width=\textwidth]{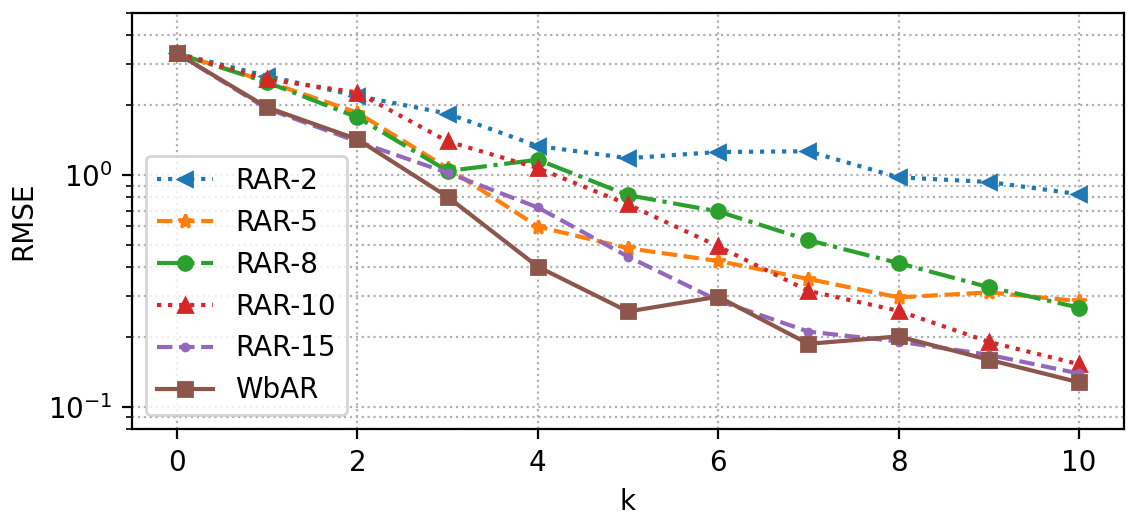}
	\caption{The RMSE of solution on the test dataset for Poisson equation comparing with RAR at $x=2,5,8,10,15$.} 
	\label{fig:poisson_comparision_rar}
    \end{subfigure}
    \hfill
    \begin{subfigure}[b]{\textwidth}
    	\centering
        \includegraphics[width=.9\textwidth]{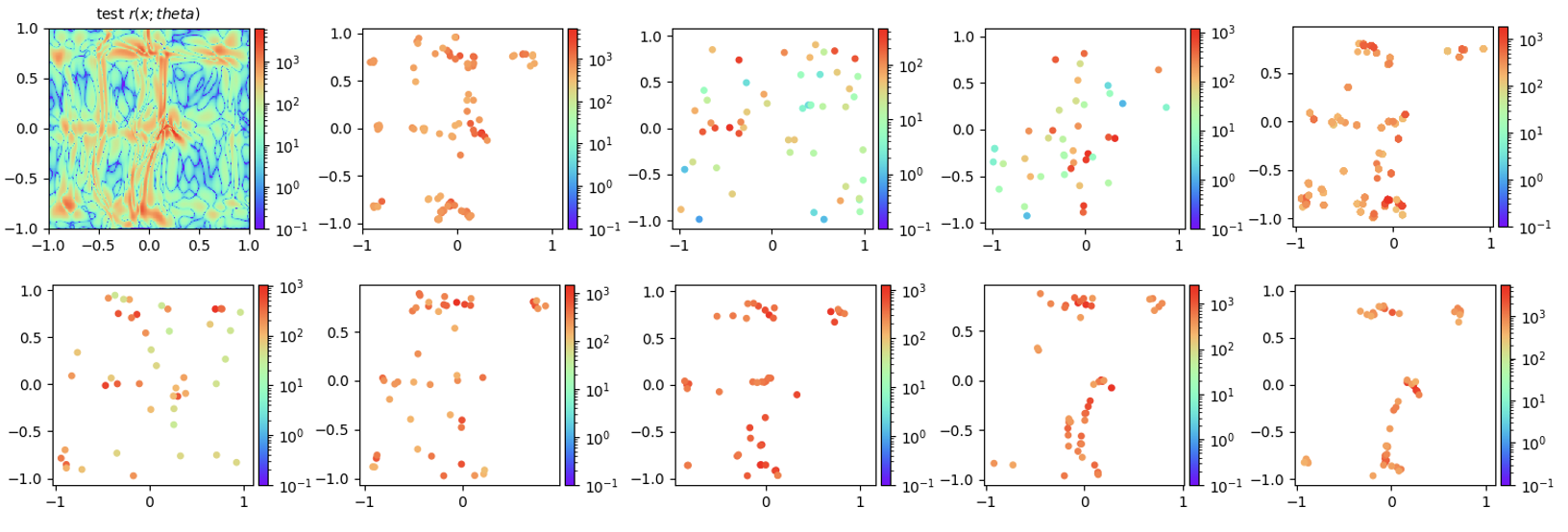}
	\caption{Samples generated by WbAR, uniform sampling, SAIS, FIPINN-R-1 (1st row from left to right), RAR-$x$, $x=2,5,8,10,15$ (2nd row from left to right) at $k=1$.}
	\label{fig:poisson_sample}
    \end{subfigure}
    \caption{Two-dimensional Poisson equation experiment results}
    \label{fig:poisson_combined}
\end{figure*}
\clearpage
}

\textit{Experiment setup:} 
The PINN model is a fully connected neural network with 8 hidden layers and 20 neurons in each layer. 
The number of boundary points ($N_b$) is set to $200$. 
The model is first trained on $500$ Latin hypercube sampling points ($N_0 = 500$) for $200,000$ epochs ($ep_0=2\text{e}5$) using Adam optimizer with a learning rate $lr=0.0001$. 
In subsequent iterations, $N_i = 100$, $ep_i=1\text{e}5$ for $i \geq 2$. 
The momentum of the Adam optimizer is reset to $0$ before each retrain. 
The detailed parameter settings for each sampling strategy are as follows:
\begin{enumerate}
\item WbAR: the number of revisiting iterations $m=1$, the maximum perturbation threshold $\epsilon=0.1$, the number of iteration steps $T=20$, and the step size of each gradient ascent iteration $\eta=0.02$. 
\item SAIS: the number of each sampling is set to $300$, the parameter $p_0=0.1$, and the maximum updated number $M=10$. 
\item RAR-$x$: we sample $x$ times of the number of samples and select the top samples maximizing the residual.
\item FIPINN-R-$x$: we sample $x \times N_0$ collocation points, with re-sampling parameters $a=0.5$ and $b=1$. 
\end{enumerate}
We use LHS as the baseline sampling method. 
The LHS baseline doesn't follow the iterative training. 
The model is trained with all the samples in one time for $2e5$ epochs. 
Models are tested on a grid that discretizes each dimension by 256.

In Figure \ref{fig:poisson_aaas}, we demonstrate the application of WbAR for Equation (\ref{eq:poisson}). 
When $k=0$, the model was trained on $N_0$ samples generated by LHS. 
The model prediction significantly differs from its true solution. 
The relative mean-squared-error (RMSE) is $3.340$. 
The residual $r(x;\theta)$ has the highest amplitude more than $6,000$ on the test dataset. 
Then, WbAR is applied to generate adversarial samples based on the $N_0$ initialization samples. 
The largest amplitude of adversarial samples exceeds $5,000$, which indicates that the adversarial attack is good at locating local maxima of the residual. 
When the training iteration reaches $k=6$, the model prediction is much closer to the true solution. 
The RMSE decreases to $0.298$. 
The distribution of the adversarial samples coincides with the residual, indicating that regions of high residual acquired more adversarial samples. 
As $k=9$, the RMSE reaches $0.160$. 
The largest residual is around $10$. 
The adversarial samples distribute similar to the distribution of the residual and the largest residual of adversarial samples is also around $9$. 
The experiment shows that, by applying WbAR to Equation (\ref{eq:poisson}), the model is converging to the true solution as the number of training iteration $k$ increases.

The performance comparisons between WbAR and other sampling strategies, including LHS, uniform sampling,  SAIS, FIPINN-R-$x$, and RAR-$x$, are shown in Figure \ref{fig:poisson_comparision} and Figure \ref{fig:poisson_comparision_rar}. 
When trained with $1,500$ LHS samples, the model achieves an RMSE of $0.699$. At iteration $k=10$, uniform sampling yields an RMSE of $1.450$ while SAIS attains $0.679$. 
FIPINN-R-1, using only $500$ collocation samples, reaches an RMSE of $1.373$ by $k=10$. For fair comparison, FIPINN-R-3 employs $1,500$ samples and achieves a slightly reduced RMSE of $1.288$.
The model trained with adversarial samples has the best performance which reaches the RMSE $0.128$ at $k=10$. 
For RAR-$x$, when more Monte Carlo samples are generated, it gains better performance. 
The RAR-15 selects $100$ top-residual samples from $1,500$ Monte Carlo samples, which achieves compatible performance with WbAR at RMSE $0.13925$.

We further plot the generated samples of all the tested sampling strategies at $k=1$. 
As shown in Figure \ref{fig:poisson_sample}, the highest residual is around $6e3$. 
The highest residual of samples generated through the uniform sampling is about $4e2$ which is far from the highest residual in the problem domain. 
The highest sample residual of SAIS is about $1e3$, however the distribution of the residual is not Gaussian-shaped. 
FIPINN-R-1 achieves a maximum residual of approximately $2e3$ while providing a reasonably accurate estimation of the failure distribution. However, the ``annealing" framework ultimately limits the algorithm's performance. (A more detailed discussion is provided in Section \ref{sec:burgers}.)
For RAR-$x$, the highest sample residual increases as $x$ increases. 
When $x=15$, the highest sample residual is $5e3$ and RAR-15 achieved the best performance among all tested RARs. 
WbAR directly locates the highest residual areas through residual gradients, thus also obtains the highest sample residual around $5e3$. 
From the sample distribution, RAR-15 inevitably locates all the samples in the highest residual regions. 
The slightly lower residual regions are ignored, for examples those on the left side of the problem domain.
Such Phenomenon is caused by the contradiction between finding the highest residual through Monte Carlo and finding all the local maximums. 
However, the mechanism of WbAR allows to find the highest residual samples in all the local maximums.

\subsection{High-dimensional Poisson equation}
\begin{figure*}[H]
    \begin{subfigure}[b]{\textwidth}
    	\centering
	\includegraphics[width=.18\textwidth]{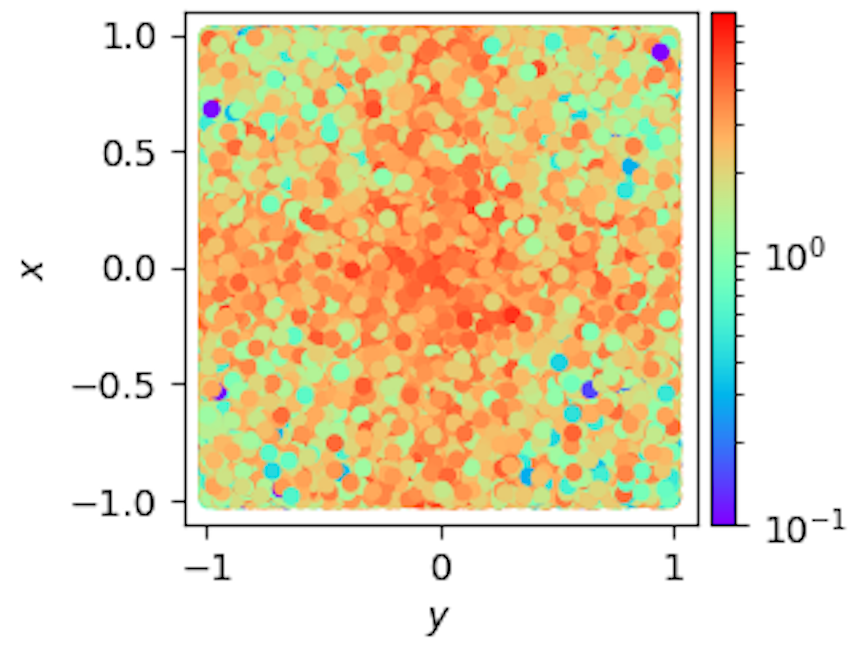}
	\caption{Projected true solution of the high-dimensional multi-peaks Poisson equation}
	\label{fig:highdim_gt}
    \end{subfigure}
    \hfill
    \begin{subfigure}[b]{\textwidth}
	\includegraphics[width=.8\textwidth]{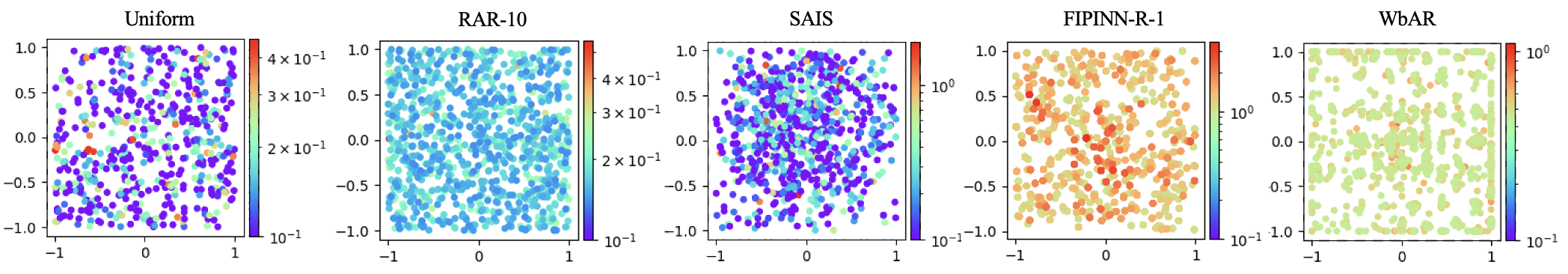}
	\caption{Samples for high-dimensional multi-peaks Poisson equation ($c=10$) at $k=3$.}
	\label{fig:highdim_c10_sample}
    \end{subfigure}
    \hfill
    \begin{subfigure}[b]{\textwidth}
        \includegraphics[width=.9\textwidth]{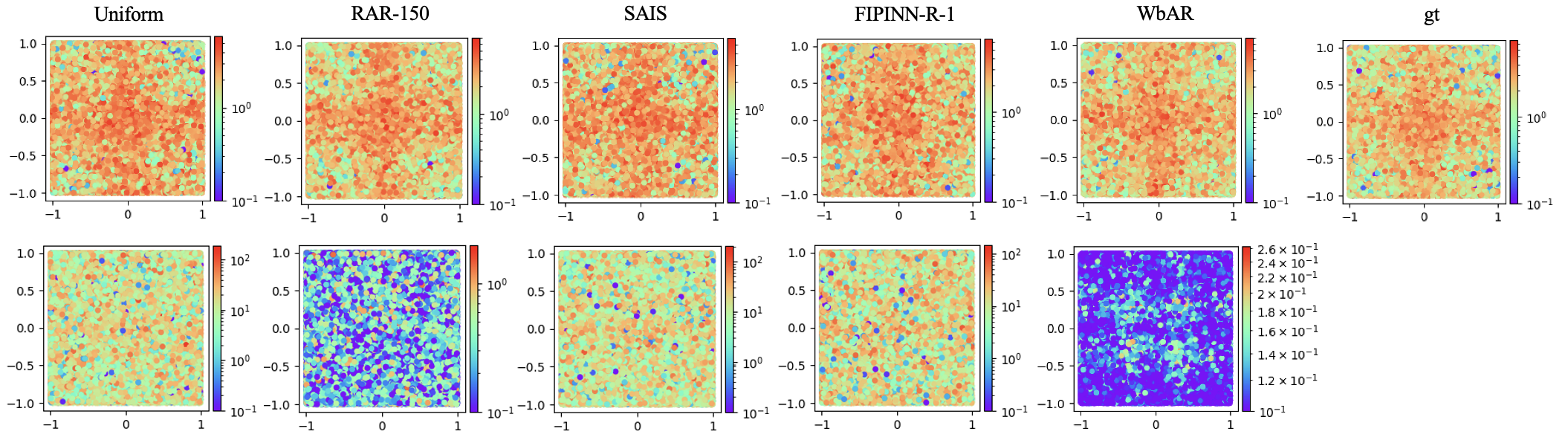}
	\caption{Model prediction and residual for high-dimensional multi-peaks Poisson equation ($c=11$) at $k=3$.}
	\label{fig:highdim_c11_sample}
    \end{subfigure}
    \hfill
    \begin{subfigure}[b]{.73\textwidth}
        \includegraphics[width=.8\textwidth]{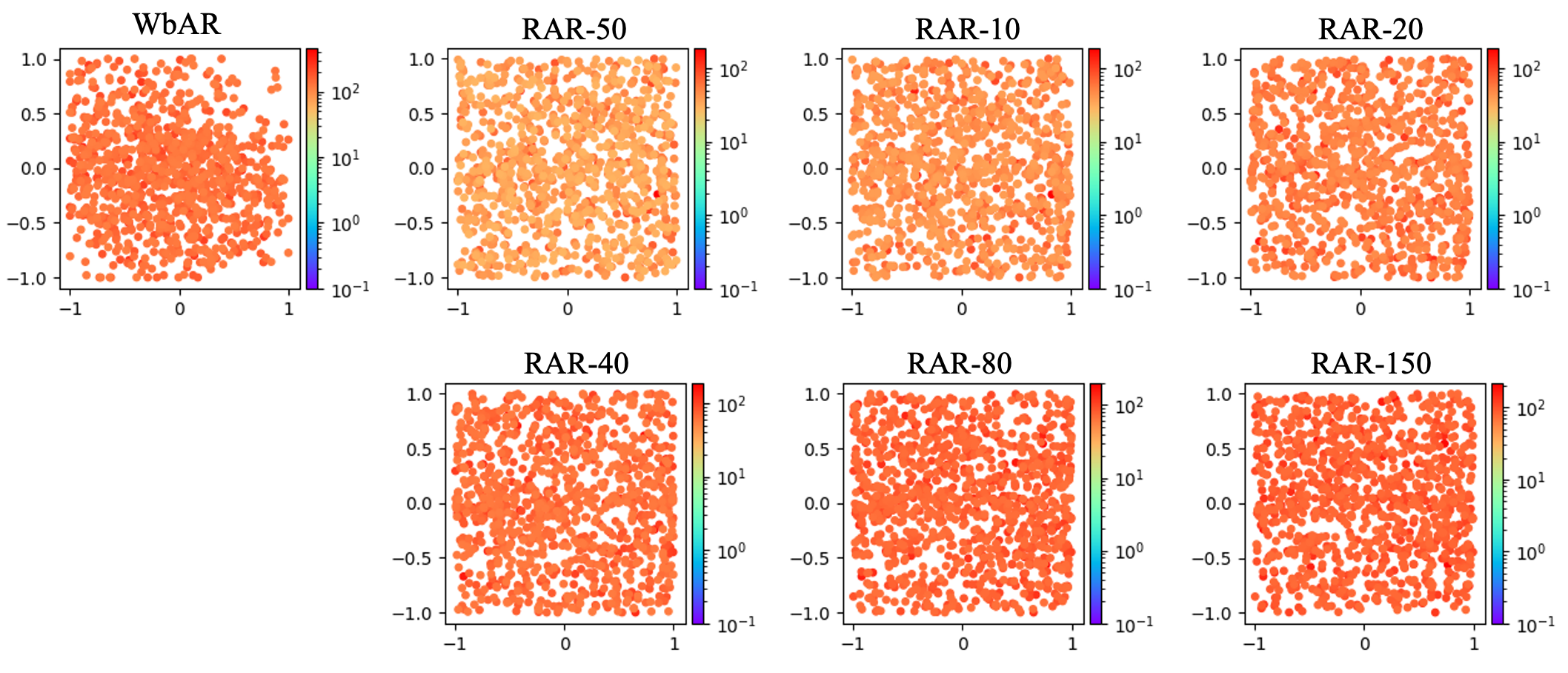}
	\caption{Samples for high-dimensional multi-peaks Poisson equation ($c=11$) at $k=1$.}
	\label{fig:highdim_c11_sample_comparision}
    \end{subfigure}
    \hfill
    \begin{subfigure}[b]{.23\textwidth}
    	\centering
        \includegraphics[width=.8\textwidth]{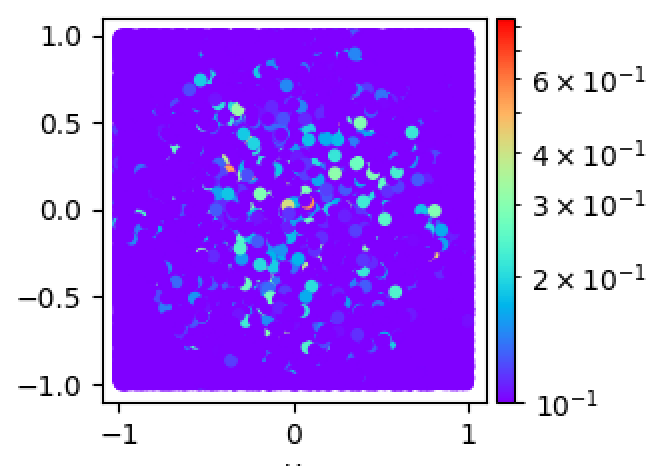}
	\caption{Projected true solution of the high-dimensional one-peak Poisson equation}
	\label{fig:highdim_1p_gt}
    \end{subfigure}
    \caption{High-dimensional Poisson equation experiment results}
    \label{fig:highdim_combined}
\end{figure*}

Then we consider the following high-dimensional Poisson equation:
\begin{align}\label{eq:poisson}
-\Delta u(x_1,\hdots,x_d) = f(x_1,\hdots,x_d) \;\;\;\;\; &\text{in } \Omega \notag \\
u(x_1,\hdots,x_d) = g(x_1,\hdots,x_d) \;\;\;\;\; &\text{on } \partial\Omega
\end{align}
where $\Omega=[-1, 1]^d$ and the solution is
\begin{align}
u(x_1,\hdots,x_d) &= \sum_{i=1}^d \text{exp}\big({-c x_i^2}\big) 
\end{align}
Samples of the true solution projected onto the first two dimensions are shown in Figure \ref{fig:highdim_gt} when $d=9$ and $c=10$.

\textit{Experiment setup:} 
The PINN model has 8 hidden layers and 40 neurons in each layer, $N_b=900$, $N_0 = 10,000$. 
The model is first trained for $100,000$ epochs with $lr=0.001$ then trained for another $100,000$ epochs with $lr=0.0001$. 
In the subsequent iterations $N_i = 1,000$. The rest parameter settings remain the same as the two-dimensional Poisson equation experiment.

\begin{table*}[!t]
\caption{Performance comparison of LHS, uniform sampling, RAR-$x$, SAIS, FIPINN-R-$1$, and WbAR for the high-dimensional multi-peaks Poisson equation ($c=10$).}
\label{tab:highdim_c10}
\centering
\scriptsize
\begin{tabular}{c |  c c c c c c c c c c} 
\hline
 RMSE & LHS & Unfrm & SAIS & FIPINN-R-1& RAR-2 & RAR-5 & RAR-10&RAR-15&RAR-20&WbAR \\
\hline
$k=0$ & 9.4e-3 &9.4e-3& 9.4e-3 & 9.4e-3 & 9.4e-3 & 9.4e-3 & 9.4e-3&9.4e-3&9.4e-3&9.4e-3\\
$k=1$ & -            &7.0e-3  & 6.2e-3 & 1.3e-2 &5.7e-3 & 5.9e-3 &4.8e-3&4.3e-3&4.3e-3&4.1e-3\\
$k=2$ & -            &5.4e-3 & 5.1e-3 & 1.0e-2& 5.0e-3  & 3.7e-3 &3.5e-3&4.1e-3&3.3e-3&4.4e-3\\
$k=3$ & 1.2e-2  &4.5e-3  & 5.0e-3& 6.3e-3 & 4.9e-3 &3.1e-3&3.0e-3&3.2e-3&3.5e-3&\textbf{2.7e-3}\\
\hline
\end{tabular}
\end{table*}

\begin{table*}[!t]
\caption{Performance comparison of LHS, uniform sampling, RAR-$x$, SAIS, FIPINN-R-$x$, and WbAR for the high-dimensional multi-peaks Poisson equation ($c=11$).}
\label{tab:highdim_c11}
\centering
\scriptsize
\begin{tabular}{c |  c c c c c c c c c c c c} 
\hline
 RMSE & RAR-2 &RAR-5&RAR-10&RAR-20&RAR-40&RAR-80&RAR-150& LHS &Unfrm&SAIS &FIPINN-R-1&WbAR \\
\hline
$k=0$ &  2.9e-1 &2.9e-1&2.9e-1&2.9e-1&2.9e-1&2.9e-1& 2.9e-1& 2.9e-1 & 2.9e-1 & 2.9e-1 &2.9e-1 & 2.9e-1\\
$k=1$ & 2.9e-1 &2.3e-1&2.3e-1&2.2e-1&2.1e-1&2.2e-1&2.1e-1& -            &2.9e-1  & 2.8e-1&2.1e-1 & 2.4e-1\\
$k=2$ & 2.8e-1 &2.2e-1&2.2e-1&2.1e-1&2.0e-1&2.1e-1&2.0e-1& -            &2.9e-1  & 2.8e-1&2.0e-1 & \textbf{4.4e-2}\\
$k=3$ & 2.8e-1 &2.1e-1&2.1e-1&2.0e-1&1.5e-1&1.8e-1&7.7e-2& 3.3e-1 &2.8e-1  & 2.8e-1& 1.6e-1& \textbf{4.4e-2}\\
\hline
\end{tabular}
\end{table*}

As demonstrated in Table \ref{tab:highdim_c10}, when $d=9$ and $c=10$, WbAR achieved the lowest RMSE after 3 iterations. 
Figure \ref{fig:highdim_c10_sample} shows that WbAR can adaptively locate failure regions in the high-dimensional space through the white-box adversarial attack. 
When $c=11$, the uniform and SAIS sampling fail to refine the model at $k=3$ while WbAR provides a boosted model performance at $k=2$ as shown in Table \ref{tab:highdim_c11} and Figure \ref{fig:highdim_c11_sample}. It can be observed that RAR-$x$ has difficulty concentrating to high residual areas in this complicated high-dimensional problem, such that a great amount Monte Carlo samples have to be taken to ensure the performance. FIPINN-R-1 demonstrates superior performance compared to both uniform sampling and SAIS methods, yet still exhibits exploration challenges in high-dimensional spaces. As shown in Figure \ref{fig:highdim_c11_sample_comparision}, WbAR finds the maximum residual around $4e2$, however RAR-150 only finds the maximum residual around $2e2$. This is because when then dimension increases, Monte Carlo sampling is facing difficulty estimating the global maximum.

Further, we consider the one-peak high-dimensional Poisson equation with the solution
\begin{align}
u(x_1,\hdots,x_d) &= \text{exp}\big({-\frac{c}{d} \sum_{i=1}^d x_i^2}\big) 
\end{align}
where $d=9$ and $c=10$ (\cite{gao2023failure}). 
Its samples of the true solution projected onto the first two dimensions are shown in Figure \ref{fig:highdim_1p_gt}. The experiment setups are kept unchanged. As shown in Table \ref{tab:highdim_1p}, the uniform sampling achieved slightly better performance comparing to the other sampling methods. Since the solution of the one-peak Poisson problem is much smoother than multi-peaks Poisson problems, the uniform sampling is more suitable for such situation.

\begin{table}[htb]
\small
\caption{Performance comparison of LHS, uniform sampling, RAR-2, SAIS, and WbAR for the high-dimensional one-peak Poisson equation.}
\label{tab:highdim_1p}
\centering
\scriptsize
\begin{tabular}{c |  c c c c c} 
\hline
 RMSE & LHS & Unfrm & RAR-2 & SAIS & WbAR \\
\hline
$k=0$ & 1.9e-2 & 1.9e-2 & 1.9e-2 & 1.9e-2 & 1.9e-2\\
$k=1$ &     -        &1.6e-2 & 1.6e-2 & 1.7e-2 &2.1e-2\\
$k=2$ &     -        &1.5e-2 & 1.5e-2 & 1.6e-2 &1.7e-2\\
$k=3$ & 1.9e-2 &\textbf{1.4e-2} & 1.4e-2 & 1.7e-2 &1.7e-2\\
\hline
\end{tabular}
\end{table}


\subsection{Burgers' Equation}\label{sec:burgers}

Consider the following Burgers' equation:
\begin{align}
&u_t + u u_x - (0.01/ \pi) u_{xx} = 0,   \notag \\
&u(x,0) = -sin(\pi x) \notag \\
&u(t,-1) = u(t,1) = 0\notag
\end{align}
where $(x,t)\in [-1,1]\times[0,1]$. 

\textit{Experiment setup:} 
All of the model settings and sampling parameters in this experiment are based on the setup of the two-dimensional Poisson equation experiment, except for $N_b=100$, $ep_0=1\text{e}5$, $ep_1=2\text{e}5$, $N_1=500$, $N_i=1,000$, for $i\geq 2$. The training epochs are adjusted to ensure sufficient training.

\begin{figure*}[H]
    \begin{subfigure}[b]{\textwidth}
    	\centering
	\includegraphics[width=.8\textwidth]{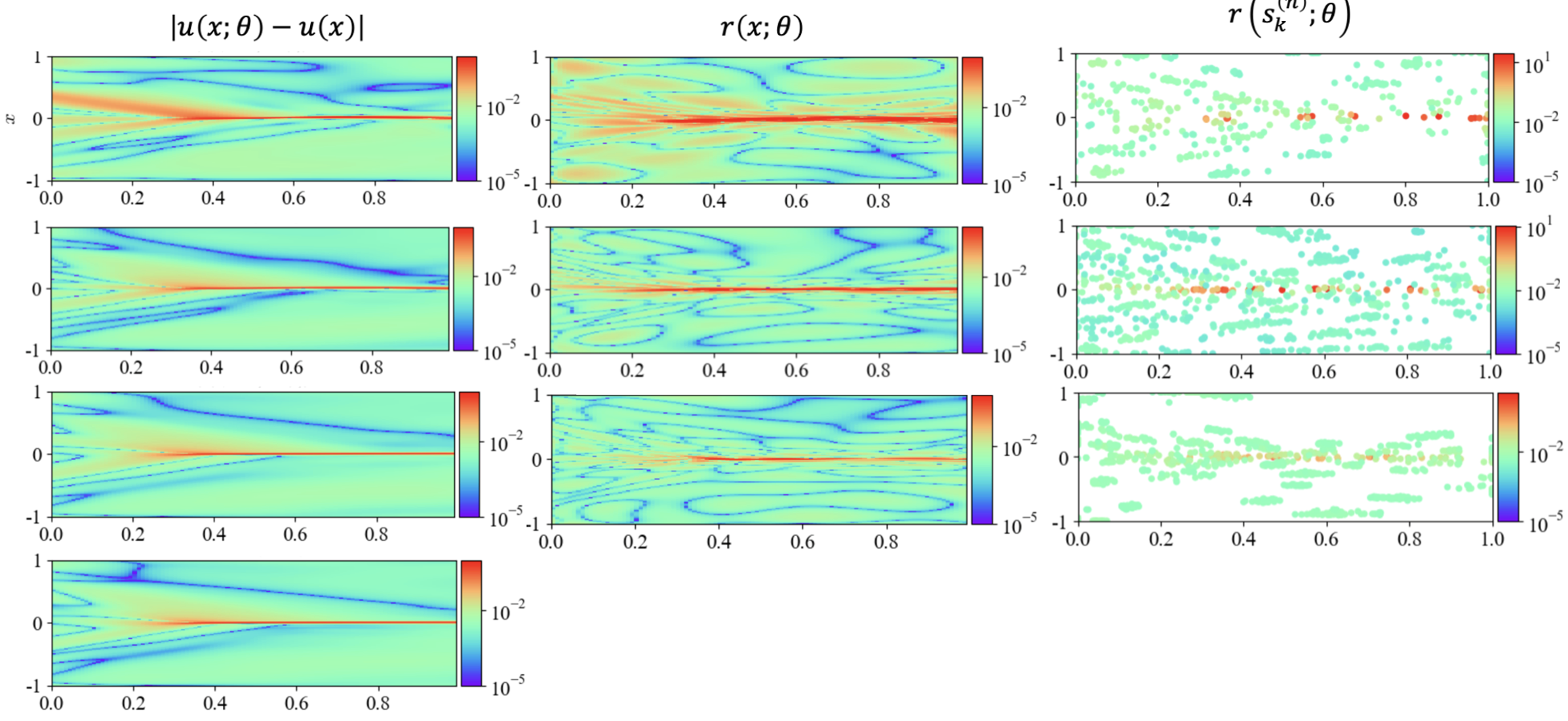}
	\caption{WbAR for the Burgers' equation at $k=0,\hdots, 3$ (1st-4th rows).}
	\label{fig:burgers_aaas}
    \end{subfigure}
    \hfill
    \begin{subfigure}[b]{\textwidth}
    	\centering
	\includegraphics[width=.9\textwidth]{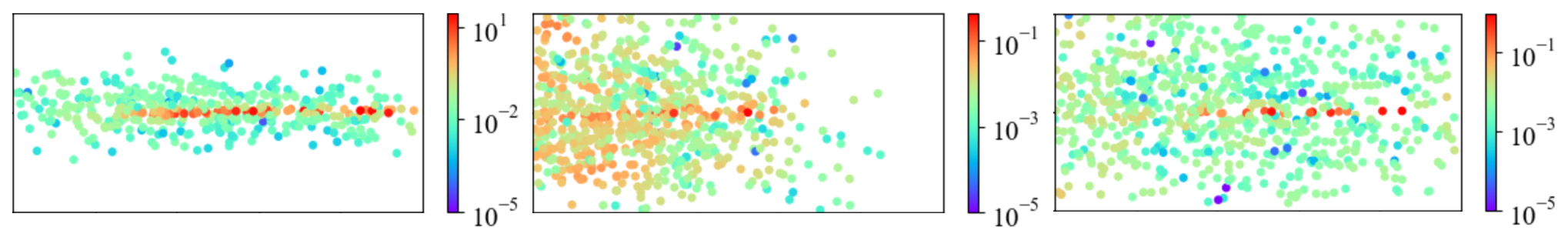}
	\caption{Samples generated by SAIS for Burgers' equation at $k=0,\hdots,2$ from left to right.}
	\label{fig:burgers_sample_sais}
    \end{subfigure}
    \hfill
    \begin{subfigure}[b]{\textwidth}
    	\centering
        \includegraphics[width=.9\textwidth]{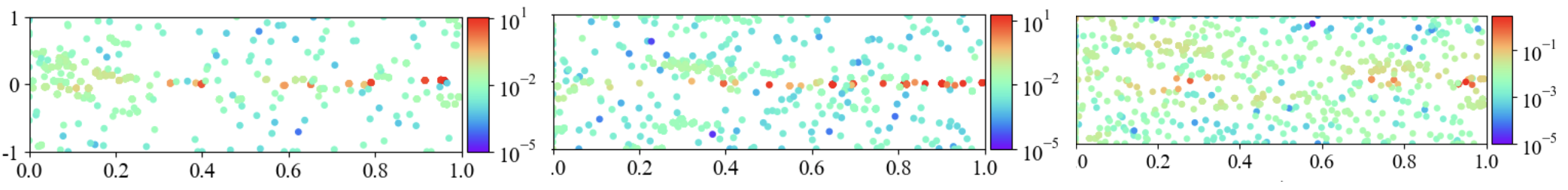}
	\caption{Samples generated by FIPINN-R-2 for Burgers' equation at $k=0,\hdots,2$ from left to right.}
	\label{fig:burgers_sample_fipinn}
    \end{subfigure}
    \hfill
    \begin{subfigure}[b]{\textwidth}
    	\centering
        \includegraphics[width=.9\textwidth]{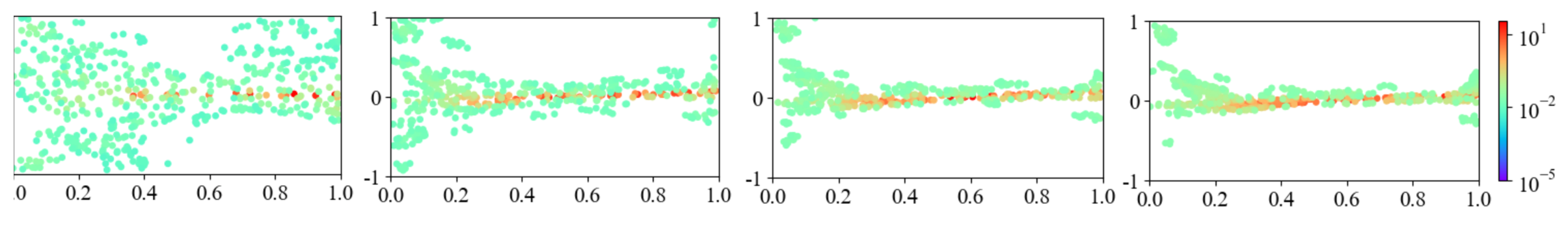}
	\caption{RAR samples at $k=1$ under $x=2,5,8,10$ from left to right.} 
	\label{fig:burgers_sample_rar}
    \end{subfigure}
    \hfill
    \begin{subfigure}[b]{.45\textwidth}
    	\centering
        \includegraphics[width=\textwidth]{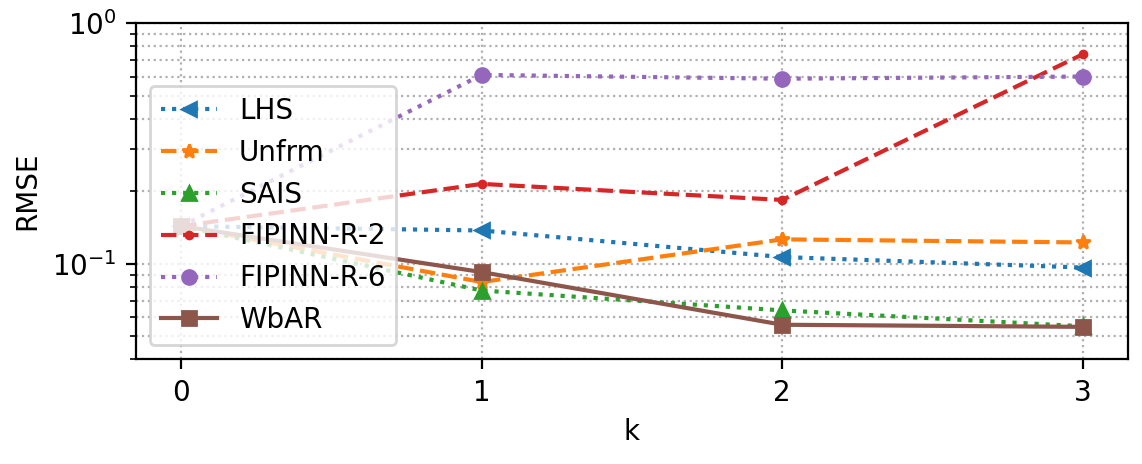}
	\caption{The RMSE of solution on the test dataset for Burgers' equation.}
	\label{fig:burgers_comparision}
    \end{subfigure}
    \hfill
    \begin{subfigure}[b]{.45\textwidth}
    	\centering
        \includegraphics[width=\textwidth]{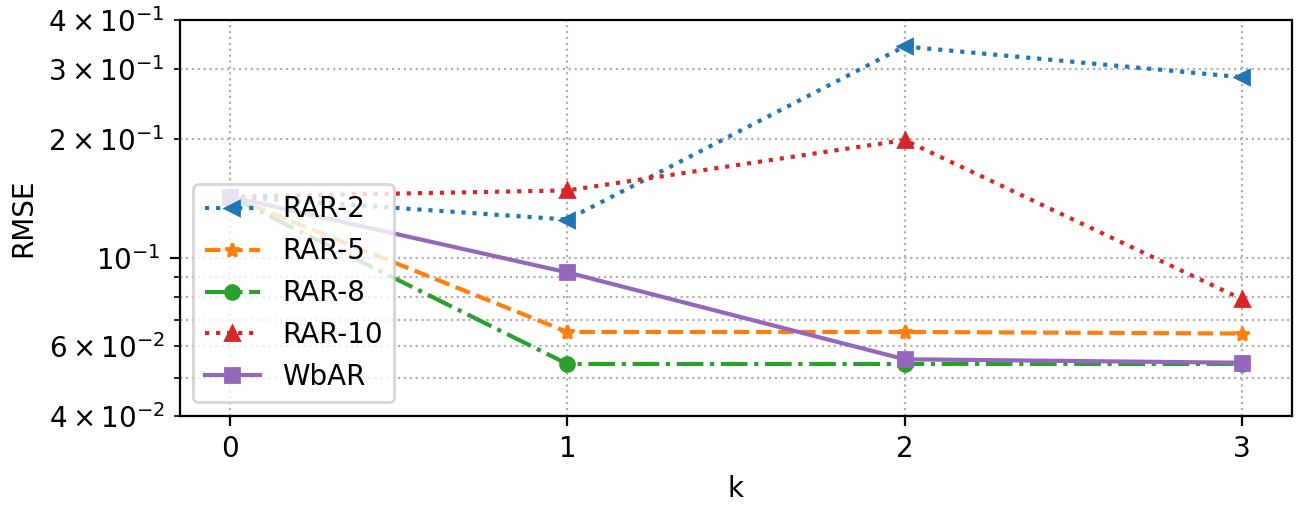}
	\caption{The RMSE of solution on the test dataset for Burgers' equation comparing with RAR at $x=2,5,8,10$.} 
	\label{fig:burgers_comparision_rar}
    \end{subfigure}
    \caption{Burgers' Equation}
    \label{fig:burgers_combined}
\end{figure*}

The training process for the Burger's equation using WbAR is demonstrated in Figure \ref{fig:burgers_aaas}. 
Initially, before the adversarial attack training iteration starts ($k=0$), the error and residual in the regions around $x=0$ and $t<0.5$ are relatively larger compared to other regions. 
WbAR adaptively generates more training samples in these regions. 
After one round of adversarial training at $k=1$, the prediction error and residual in the region $t<0.5$ are successfully suppressed. 
The regions with large error and residual around $x=0$ are significantly reduced. 
As the training progresses to $k=2$ and $k=3$, the error and residual of the model are further decreased, and the generated samples gradually gather towards the remaining region with high residual.

In approximating the Burgers' equation, high-residuals are predominantly located at $x=0$. 
As the iterative training progresses and the high-residual region diminishes, the probability of uniformly generated samples falling into this region decreases. 
SAIS employs Gaussian distribution re-estimations during the iterative training process, as shown in Figure \ref{fig:burgers_sample_sais}. 
However, the distribution of the high-residual region is not a typical Gaussian. 

As illustrated in Figure \ref{fig:burgers_sample_fipinn}, FIPINN-R-2 utilizes subset simulation to achieve a reasonably accurate estimation of the failure region distribution. However, the ``annealing" framework's first training iteration employs the maximum number of failure region samples, which causes the model to become oversaturated with these samples. This oversaturation effectively erases knowledge gained from prior iterations (a "forgetting" phenomenon). Although progressively reducing the number of failure region samples in subsequent iterations partially alleviates this issue, it concurrently reduces the model's ability to further improve the failure region characterization. These effects can also be found in the RMSE plots from the numerical experiments reported in \cite{GTYZ23}.

On the other hand, RAR selects the top-residual samples to refine the sampling towards high-residual regions. As shown in Figure \ref{fig:burgers_sample_rar}, when $x$ is too small (for example $x=2$), RAR finds difficulty to locate the high-residual area. 
When $x$ is too large (for example $x=10$), RAR concentrates all samples to the high-residual area and may break the sample balance. 
A good training strategy for RAR would be adding small amount of data for each iteration, but it significantly increases the training cost. 

For WbAR, the adversarial attack finds the local maximum of the model residual, effectively moving the samples towards the high-residual regions. The adversarial samples generated during the iterative training can be seen as random walks. The random walk ensures that new samples remain related to the samples generated in the previous iteration, which were located in high-residual regions in the previous iteration. Moreover, the revisiting mechanism, similar to RAR, further reinforces the selection of high-residual samples. 

During the PINN training, WbAR achieves the lowest RMSE ($0.0545$) of the solution at $k=2$. 
The SAIS method also achieves a similar RMSE, but with $1,000$ additional samples at $k=3$ (Figure \ref{fig:burgers_comparision}). 
By using uniform sampling and LHS, the RMSE is reduced to $0.122$ and $0.0963$ at $k=3$, respectively. 
FIPINN-R-2 achieves an RMSE of $0.743$ using $1,000$ collocation point, which is notably higher than the RMSE obtained after initial training. FIPINN-R-6 employs $3,000$ collocation points, matching the sample count used by WbAR, RAR, and SAIS at the third iteration ($k=3$), yet fails to demonstrate better performance.

For a carefully selected $x=8$, RAR reaches the lowest RMSE at $k=1$, which is the fastest. 
As shown in Figure \ref{fig:burgers_comparision_rar}, intuitively $x=5,8,10$ are descent refinement sampling, however $x=8$ has much better performance than $x=5,10$. 
It is worth mentioning that the lowest RMSE obtained for the same structure on Burgers' equation is around $0.05$, as verified in \cite{gao2023failure}. 
Thus, the rest iterations are not shown in Figure \ref{fig:burgers_comparision}. 

\subsection{A Multiscale Problem}
It has been found that the vanilla PINN fails to solve multi-scale problems \cite{leung2022nh}. For example, consider the following elliptic equation:
\begin{align}\label{eq:multiscale}
-\nabla \cdot (\kappa \nabla u(x)) = f& \;\;\;\;\; \text{in }\Omega \notag \\
u(x)=0& \;\;\;\;\; \text{on } \partial \Omega
\end{align}
where $\Omega=[0, \pi]$ and $f=sin(x)$. 
If $\kappa(x)$ has multi-scale properties, for example
$\kappa(x) = 0.5sin(2\pi x/\varepsilon) + sin(x) + 2$, where $\varepsilon=1/4$, PINN cannot converge to a satisfied solution. 
In this experiment, we demonstrates that by employing proper sampling strategies, the model can effectively approximate the multi-scale problem.

\begin{figure*}[H]
    \begin{subfigure}[b]{\textwidth}
    	\centering
	\includegraphics[width=.795\textwidth]{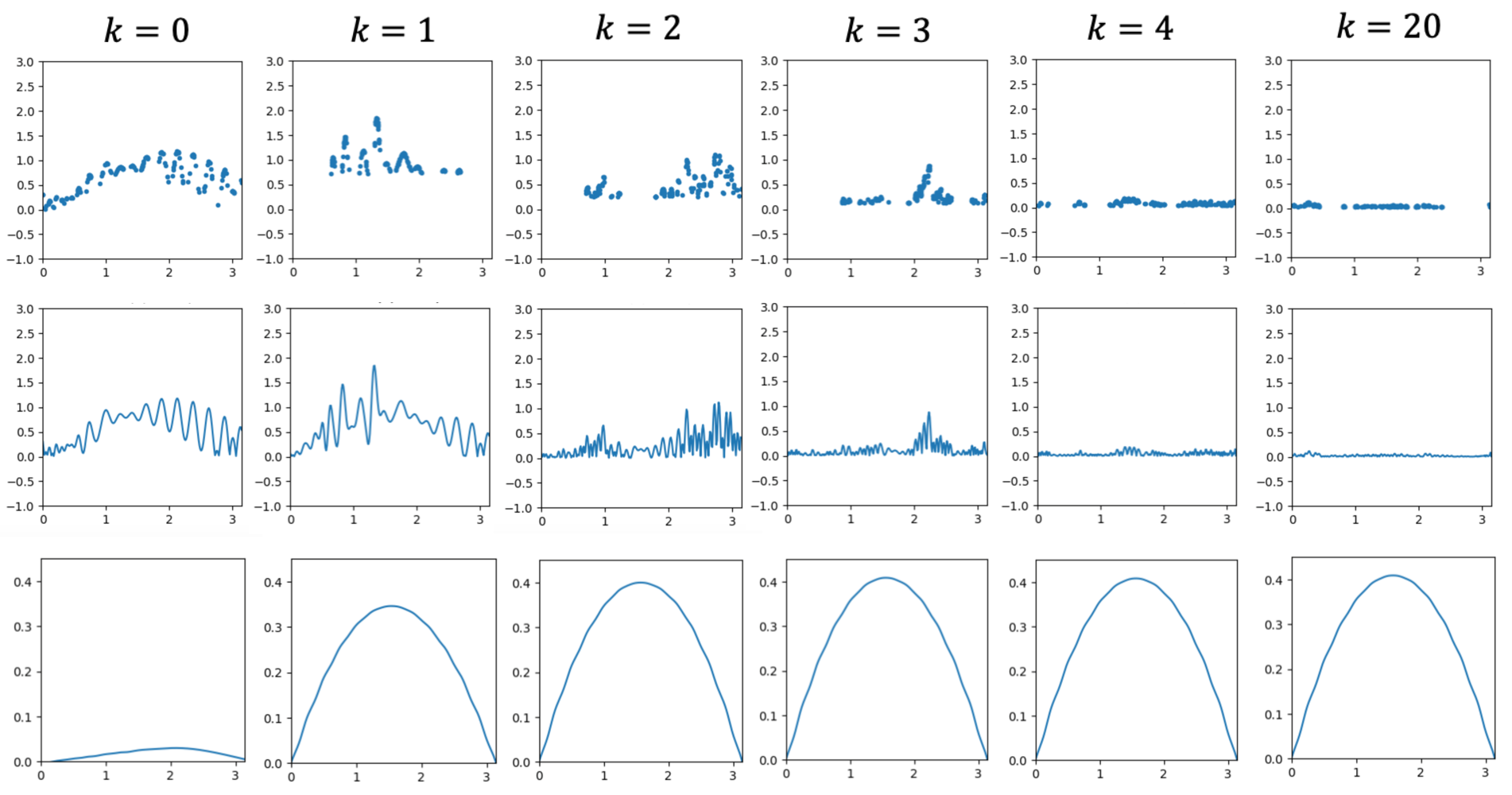}
	\caption{WbAR for the multi-scale elliptic equation. (1st row) adversarial samples; (2nd row) residual; (3rd row) prediction.}
	\label{fig:multiscale_aaas}
    \end{subfigure}
    \hfill
    \begin{subfigure}[b]{\textwidth}
    	\centering
	\includegraphics[width=.8\textwidth]{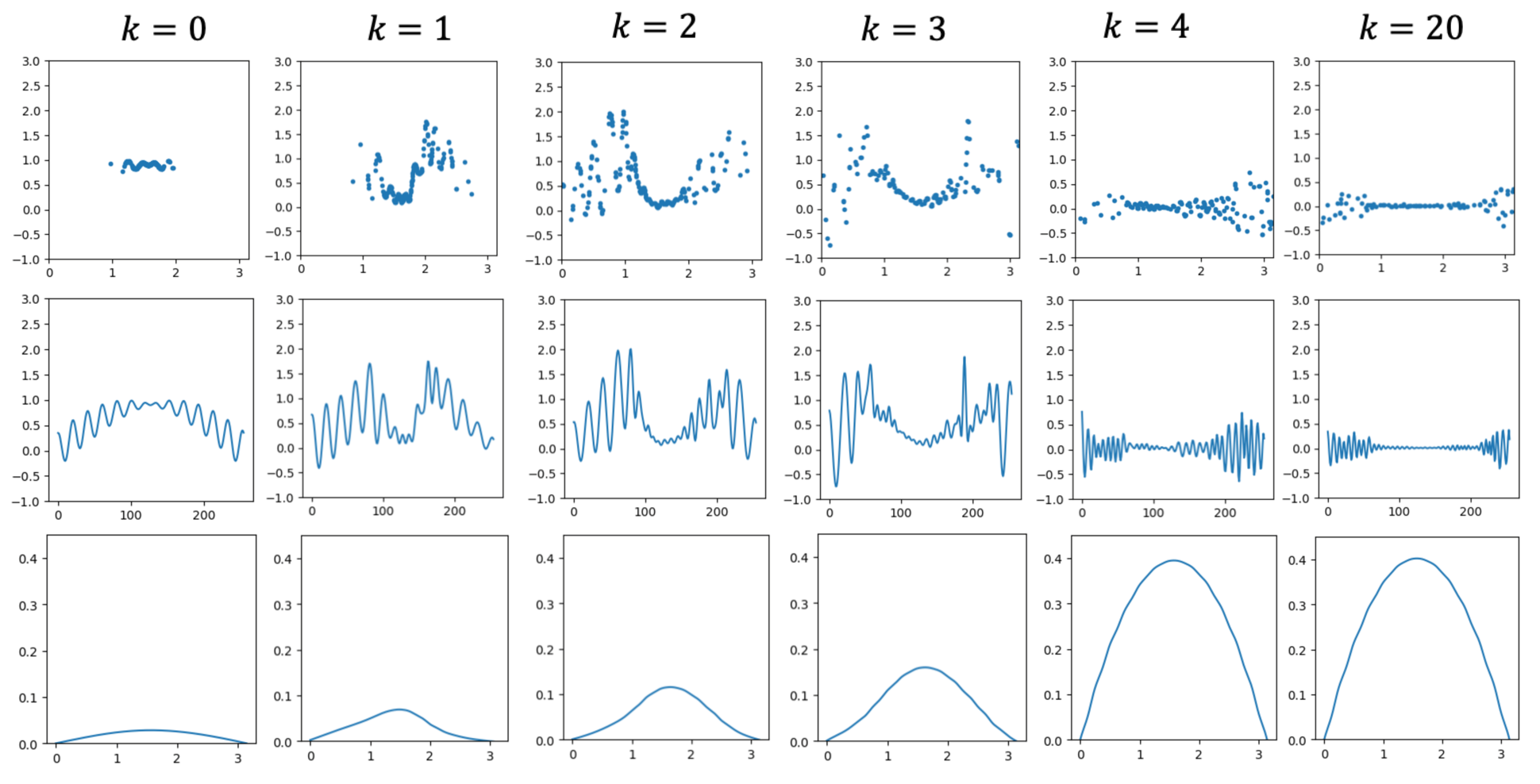}
	\caption{SAIS for the multi-scale elliptic equation.}
	\label{fig:multiscale_sais}
    \end{subfigure}
    \hfill
    \begin{subfigure}[b]{\textwidth}
    	\centering
        \includegraphics[width=.8\textwidth]{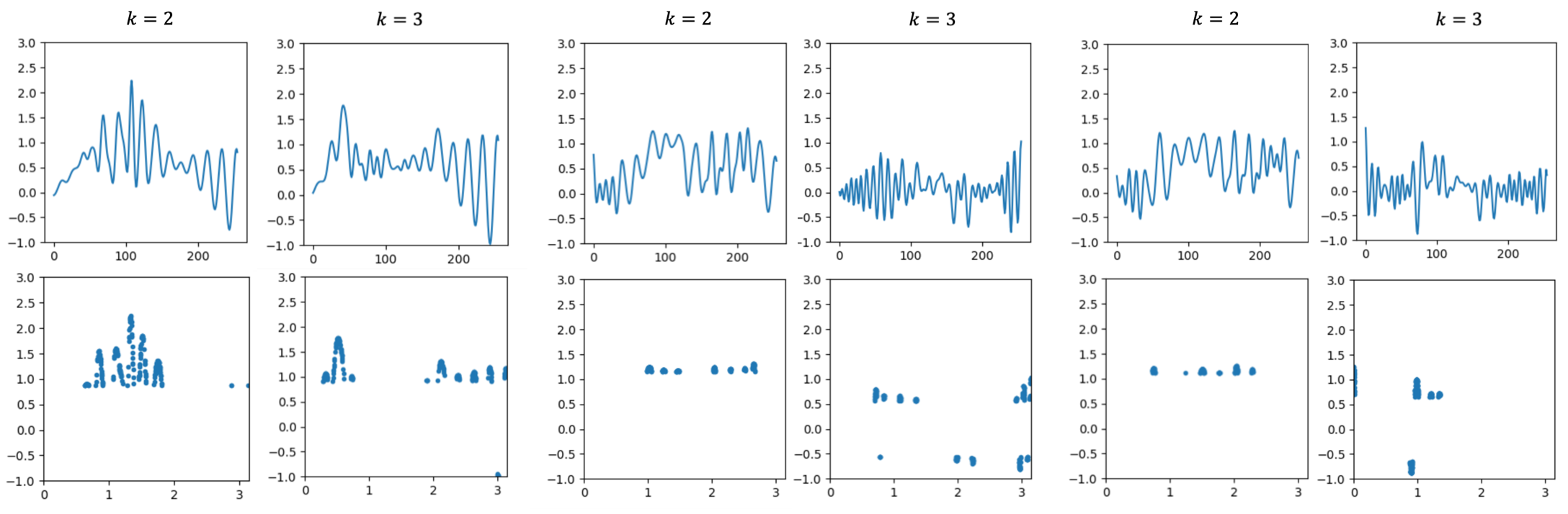}
	\caption{RAR-$x$ for the multi-scale elliptic equation (left: $x=5$, middle: $x=10$, right: $x=15$).}
	\label{fig:multiscale_rar}
    \end{subfigure}
    \caption{Multiscale problem experiment results}
    \label{fig:multiscale_combined}
\end{figure*}

\textit{Experiment setup:} 
The PINN model structure, optimizer, and learning rate settings follow the setup of the two-dimensional Poisson equation experiment. 
The sampling and training parameters are as follows: $N_i=200$, $ep_i=3e4$ for $i\geq 0$. 
For the WbAR method, $m=2$, $\epsilon=0.2$, and $T=2$.
The remaining parameter settings for WbAR and other sampling strategies are kept unchanged. 
A scale of $200$ is multiplied to the boundary loss for all sampling strategies to balance the effect of having fewer boundary sampling points compared to collocation sampling points in one-dimensional problems. 
The adversarial attack parameters are adjusted in order to allow the model to explore the problem domain adequately before focusing on local maxima. 

In Figure \ref{fig:multiscale_aaas}, we demonstrate WbAR for the problem (\ref{eq:multiscale}). 
The residual shows high-frequency oscillations and WbAR generates samples at almost each residual local maximums.
From $k=1$ to $k=4$, the model gradually converges to the true solution. 
Although the residual still contains high-frequency components, its oscillations amplitude is compressed within a much smaller range.  
When we keep increasing $k$, the model can successfully approximate the equation.  

\begin{figure}[h]
        \begin{subfigure}[b]{.45\textwidth}
    	\centering
        \includegraphics[width=\textwidth]{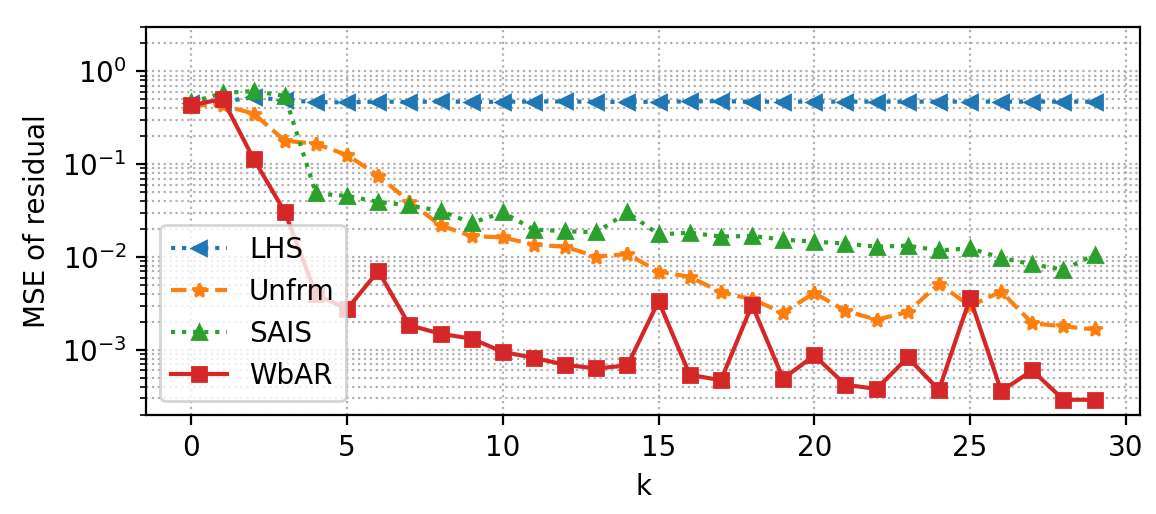}
	\caption{The MSE of the residual on the test dataset for the multi-scale problem.}
	\label{fig:multiscale_comparision}
    \end{subfigure}

    \begin{subfigure}[b]{.45\textwidth}
    	\centering
        \includegraphics[width=\textwidth]{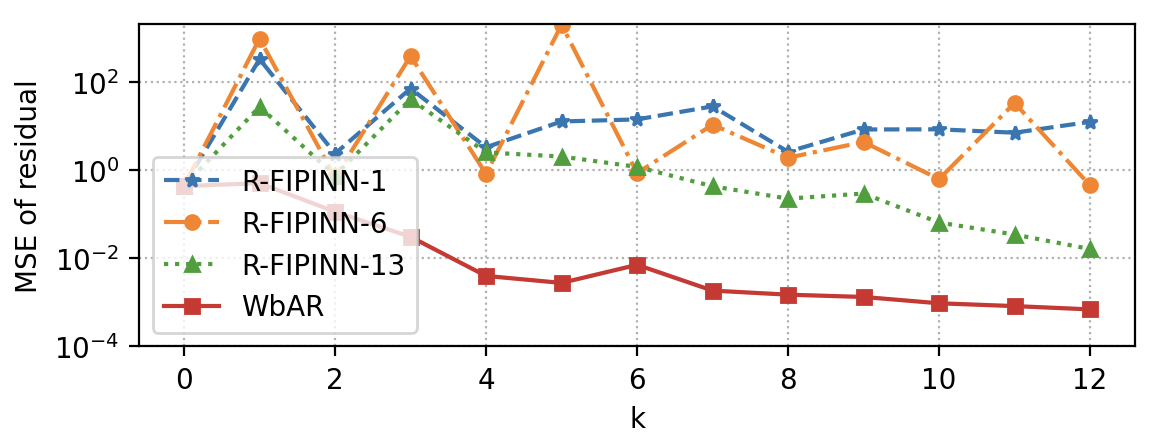}
	\caption{The MSE of the residual on the test dataset for the multi-scale problem through FIPINN-R-$x$ with $x=1,6,13$.}
	\label{fig:multiscale_comparision_fipinn}
    \end{subfigure}

    \begin{subfigure}[b]{.45\textwidth}
    	\centering
        \includegraphics[width=\textwidth]{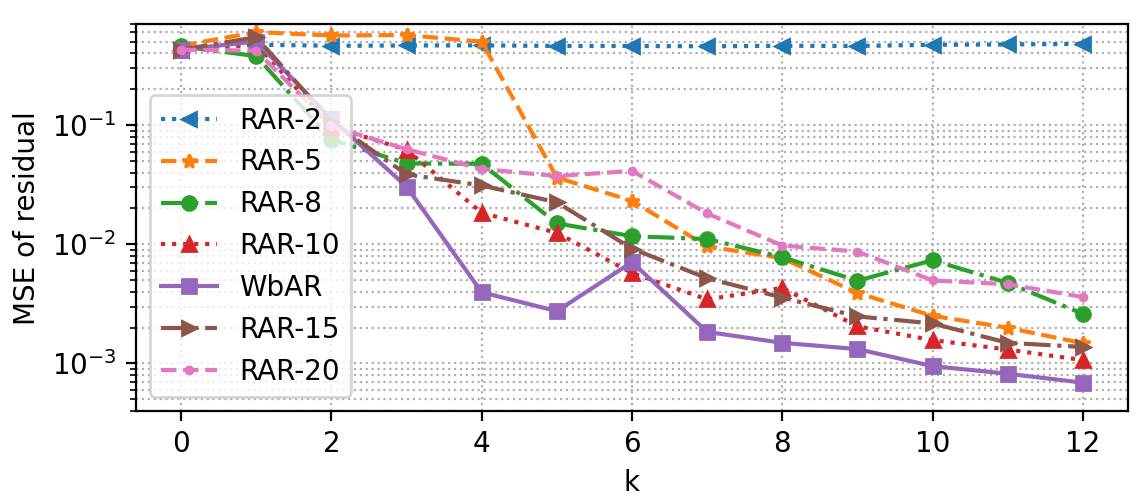}
	\caption{The MSE of the residual on the test dataset for the multi-scale problem through RAR-$x$ with $x=2,5,8,10,15,20$.}
	\label{fig:multiscale_comparision_rar}
    \end{subfigure}
    \caption{Multiscale problem algorithm comparisons}
    \label{fig:multiscale_comparison_combined}
\end{figure}

Comparing to WbAR, SAIS can also approximate the equation (\ref{eq:multiscale}) to some extent. 
As shown in Figure \ref{fig:multiscale_sais}, when $k=0$, the variance of the residual is small and the samples are concentrated in a small region. 
As $k$ continues to increase, the approximation accuracy improves.
However, most samples are generated in the central region instead of around the initial values. 
This is because of the mismatch between the residual distribution and the artificially assumed distribution. 
It results slower convergence and larger error comparing to WbAR, which is more obvious in Figure \ref{fig:multiscale_comparision}. 
Compared to uniform sampling (Figure \ref{fig:multiscale_comparision}) and FIPINN-R (Figure \ref{fig:multiscale_comparision_fipinn}), WbAR also demonstrates superior performance.

In Figure \ref{fig:multiscale_rar}, we demonstrate the samples generated by RAR-$x$ at $x=5,8,10$. 
It can be observed that RARs ignore local maximums but focuses on searching the global maximum through Monte Carlo. 
As $x$ increases, more local maximums are ignored. 
As shown in Figure \ref{fig:multiscale_rar},  RAR-$x$ achieves the best performance at $x=10$. 
When $x=15$ or $x=20$ the performance decreases, thus there is balance between finding the global residual maximum while maintaining local residual maximums.

We compare WbAR with RARs under $x=2,5,8,10,15$, $20$. As the number of Monte Carlo sampling increases, the performance of RAR increases, but the sampling cost increases as well. As shown in Figure \ref{fig:multiscale_comparision_rar}, the carefully chosen $x=10$ achieves the best performance for RAR. Higher or lower $x$ will result in degraded performance. 

\subsection{Diffusion Equation}
To make a fair comparison with RAR, we applied WbAR to the one-dimensional diffusion equation under the same setting with \cite{WZTKL23}. Consider the diffusion equation:
\begin{align}
&\frac{\partial u}{\partial t} = \frac{\partial^2 u}{\partial x^2} + e^{-t}(-\sin(\pi x)+\pi^2\sin(\pi x)),\notag\\
&x\in[-1,1], t\in[0,1] \notag\\
&u(x,0)=\sin(\pi x) \notag\\
&u(-1,t) = u(1,t)=0
\end{align}
The exact solution is $u(x,t)=\sin(\pi x)e^{-t}$.

\textit{Experiment setup:} 
The experiment setup strictly follows [1]. Specifically, the model has 4 hidden layers and 32 neurons in each hidden layer. The model is first trained with 10 residual points for $1e4$ steps. Afterwards, one more point is sampled and added for each $1000$ steps. The optimizer is Adam and the learning rate is $1e-3$. The hyper-parameters of WbAR are: $N_i=1$, $m=1$, $\epsilon=1$, $\eta=0.1$, $T=8$. The Monte-Carlo sampling for RAR is $1e5$. We tested RAR and WbAR for 8 runs and compared their average relative L2 error.

\begin{figure}[hbt]
\centering
\includegraphics[width=.45\textwidth]{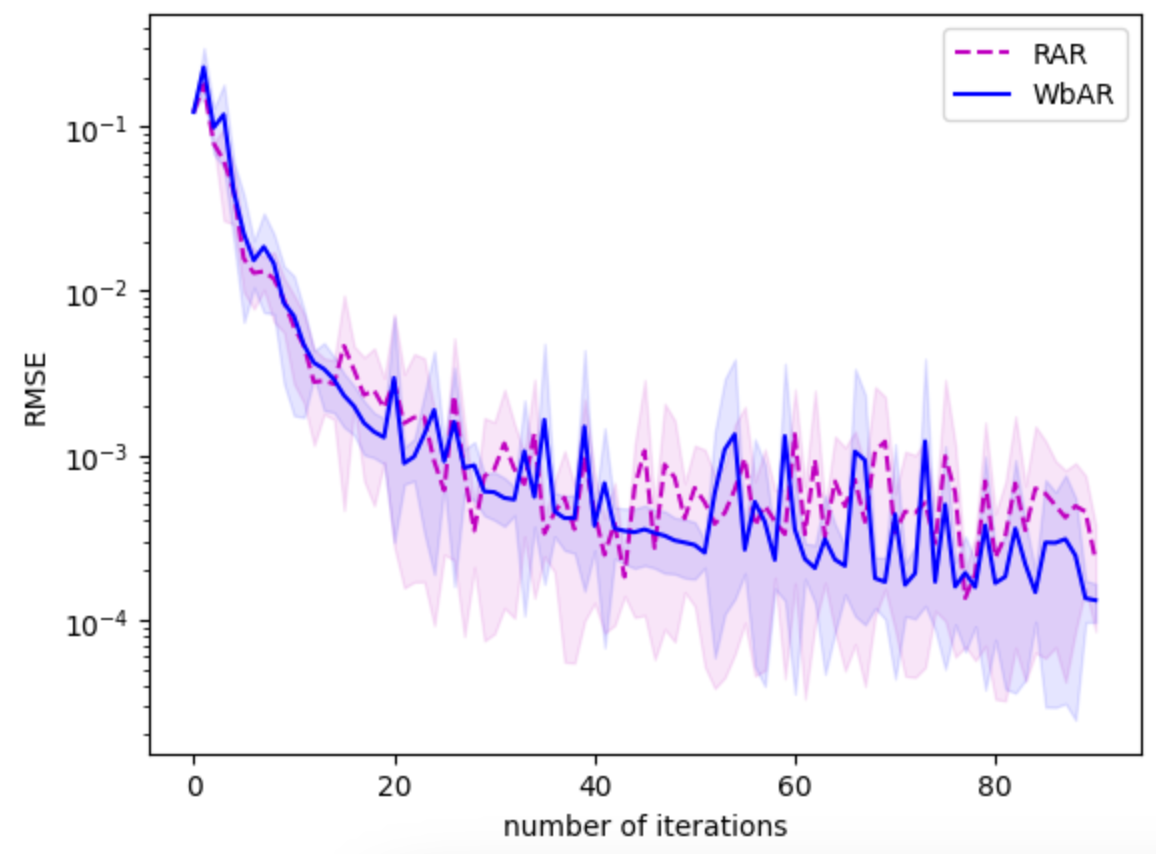}
\caption{The RMSE of solution on the test dataset for the one-dimensional diffusion equation.}
\label{fig:fair_test}
\end{figure}

As shown in Figure \ref{fig:fair_test}, WbAR has lower RMSE than RAR. The time cost of the sampling is $6.7e-3$ seconds for RAR and $9.1e-3$ for WbAR. Although WbAR has higher time cost than RAR in such one-dimensional problem, in high-dimensional problems, Monte-Carlo sampling will be less effective as we demonstrated in the high-dimension Poisson problem. The training time cost for each sampling is $1.52$ seconds, which is $200$ times larger than the sampling. Thus, comparing to the training time, the sampling time can be ignored.

\subsection{Allen-Cahn Equation}

There are also reports that PINN may violate physical causation \cite{wang2022respecting}, for example the Allen-Cahn equation in the following form:
\begin{align}
&u_t = D u_{xx} + 5(u-u^3)\notag \\
&u(x,0) = x^2\cos(\pi x) \notag \\
&u(-1,t)=u(1,t)=0
\end{align}
where $(x,t)\in[-1,1]\times [0,1]$ and $D=0.0001$. While solving the Alen-Cahn equation with PINNs, the residual is large near the initial state and decays to nearly zero after t = 0.5. We yield that the PINN fails because it accidentally converged to a local minimum and the optimizer are not capable of dragging the model out. The PINN approximation has a large flatten area around $u=0$, as $u=0$, $u_{xx}=0$, $u_t=0$ is actually a zero point for the residual $u_t - D u_{xx} - 5(u-u^3)$.

\begin{figure*}[h]
        \begin{subfigure}[b]{\textwidth}
    	\centering
        \includegraphics[width=.74\textwidth]{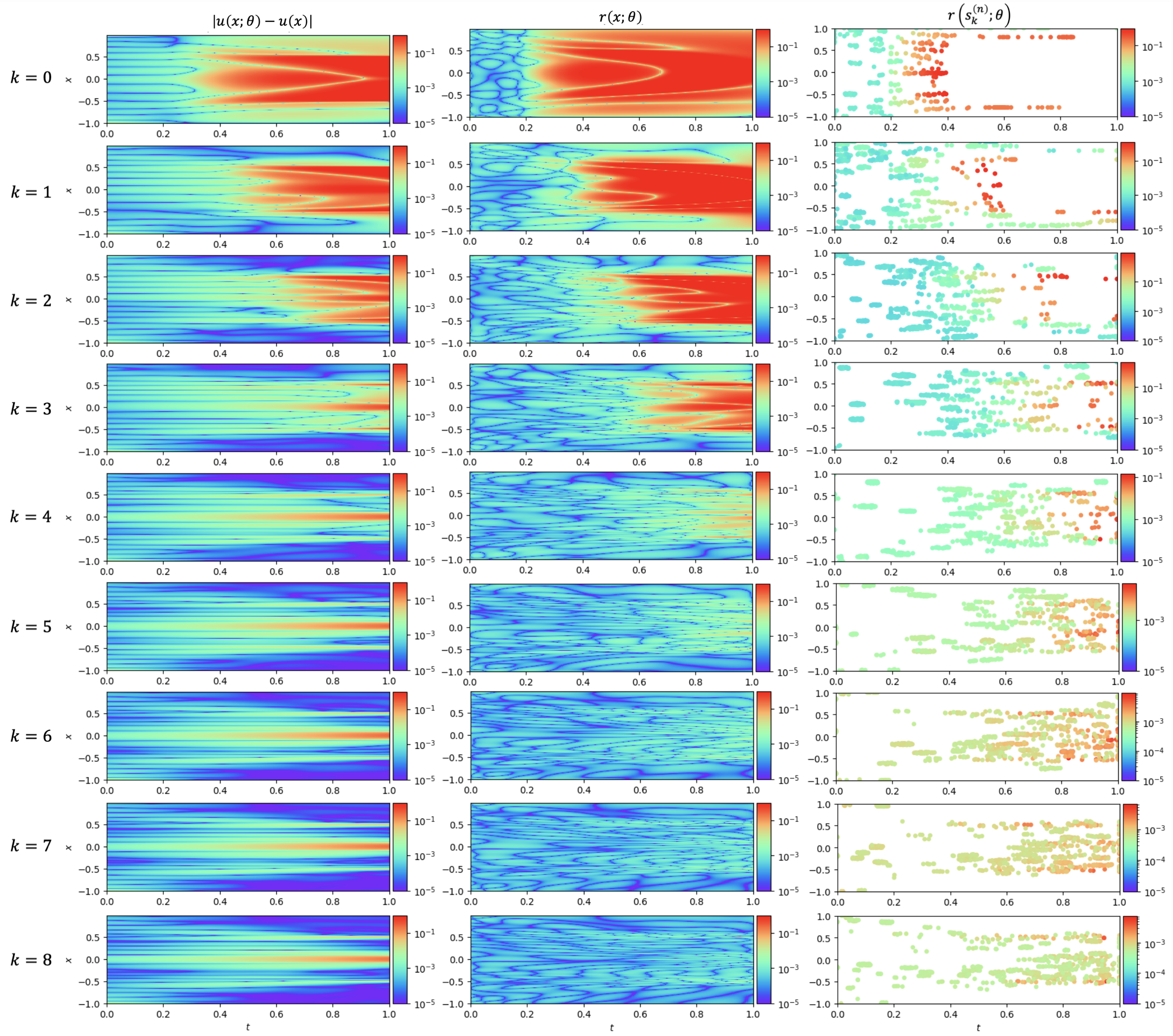}
	\caption{The adaptive temporal causal inference of WbAR for the Allen-Cahn equation.}
	\label{fig:allen_cahn_aaas}
    \end{subfigure}
    
    \begin{subfigure}[b]{.34\textwidth}
    	\centering
        \includegraphics[width=\textwidth]{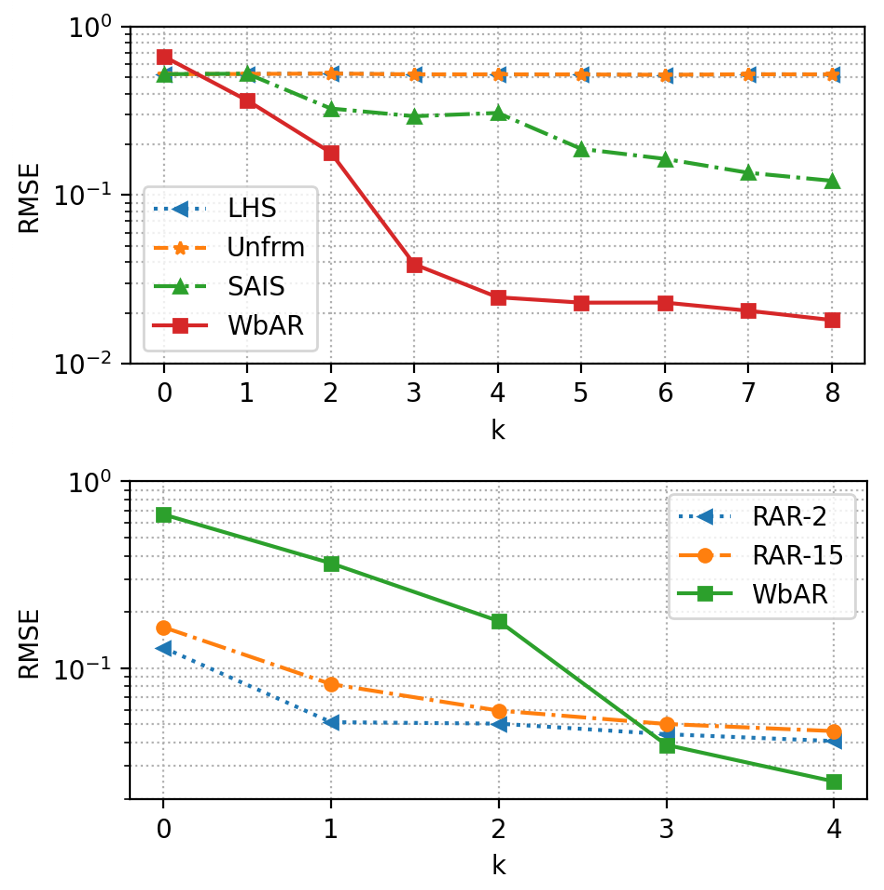}
\caption{The RMSE of solution on the test dataset for the Allen-Cahn equation.}
\label{fig:allen_cahn_comparision}
    \end{subfigure}
    \hfill
    \begin{subfigure}[b]{.63\textwidth}
    	\centering
        \includegraphics[width=.9\textwidth]{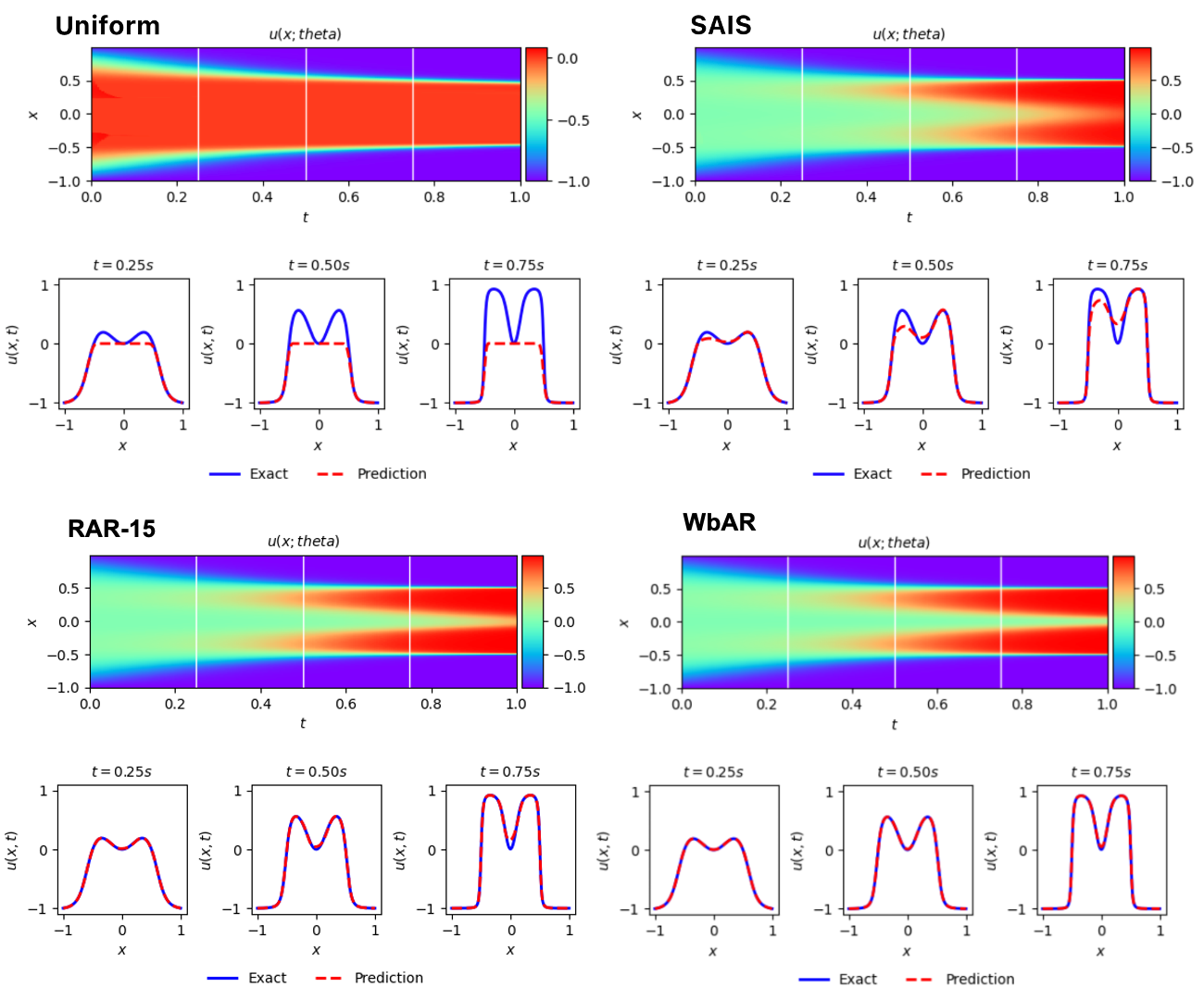}
	\caption{Uniform sampling and SAIS for the Allen Cahn equation approximation ($k=8$).}
	\label{fig:allen_cahn_converge}
    \end{subfigure}
    \caption{Allen-Cahn Equation experiment results}
    \label{fig:allen_cahn_combined}
\end{figure*}

As shown in Figure \ref{fig:allen_cahn_aaas}, we first train the PINN on $500$ collocation points within $0\leq t \leq 0.2$ sampled by LHS. The adversarial attack try to find the failure regions based on the current samping points, thus the adversarial samples move along the temporal direction adaptively (as shown in the 3rd plot in the 1st row). As the iterative training processes, some adversarial samples adaptively keep moving towards the temporal direction while the rest adversarial samples reinforce the earlier temporal region as shown at $k=1$ and $k=2$. At $k=4$, adversarial samples can cover the entire temporal interval. The solution error and the residual are significantly reduced. Start from $k=5$, adversarial samples start to focus on the two ridges and the canyon between the ridges which are the failure regions of the PINN. It can be observed from solution error, the failure regions gradually shrink as more adversarial samples are generated in the failure regions. 

\textit{Experiment setup:} All of the model settings and sampling parameters in this experiment are based on the setup of the two-dimensional Poisson equation experiment, with the exception of $N_b=400$,  $N_i=500$, $ep_i=1\text{e}6$ for $i\geq 0$. The training epochs are adjusted to ensure sufficient training. Additionally, we set $m=1.5$ to slow down the speed of the temporal inference. Specially, we only sample collocation points within $0\leq t \leq 0.2$ at $k=0$ as described above. The maximum perturbation threshold is set to $\epsilon=0.2$ such that with four training iterations in maximum, adversarial samples are capable of covering the entire temporal interval. 

In Figure \ref{fig:allen_cahn_comparision}, the adaptive behavior of adversarial training is more clearly observed. For $k<4$, the RMSE decreases rapidly because the adversarial samples primarily facilitate temporal inference. When $k\geq 4$, the adversarial samples shift their focus to refining failure regions. In contrast, LHS, RAR, and uniform sampling fail to converge to the true solution. SAIS and RAR-$x$ enables the PINN to gradually converge to the true solution (Figure \ref{fig:allen_cahn_converge}).

\subsection{Hyper-parameter Studies for WbAR} \label{sec:hyperparam}
We demonstrated the performance and adaptive behavior of the proposed WbAR on multi-peak, sharp or oscillatory solutions. 
In this section, we will study the effect of WbAR hyper-parameters on the multi-scale problem.

\begin{figure*}[h]
    \begin{subfigure}[b]{.45\textwidth}
    	\centering
        \includegraphics[width=\textwidth]{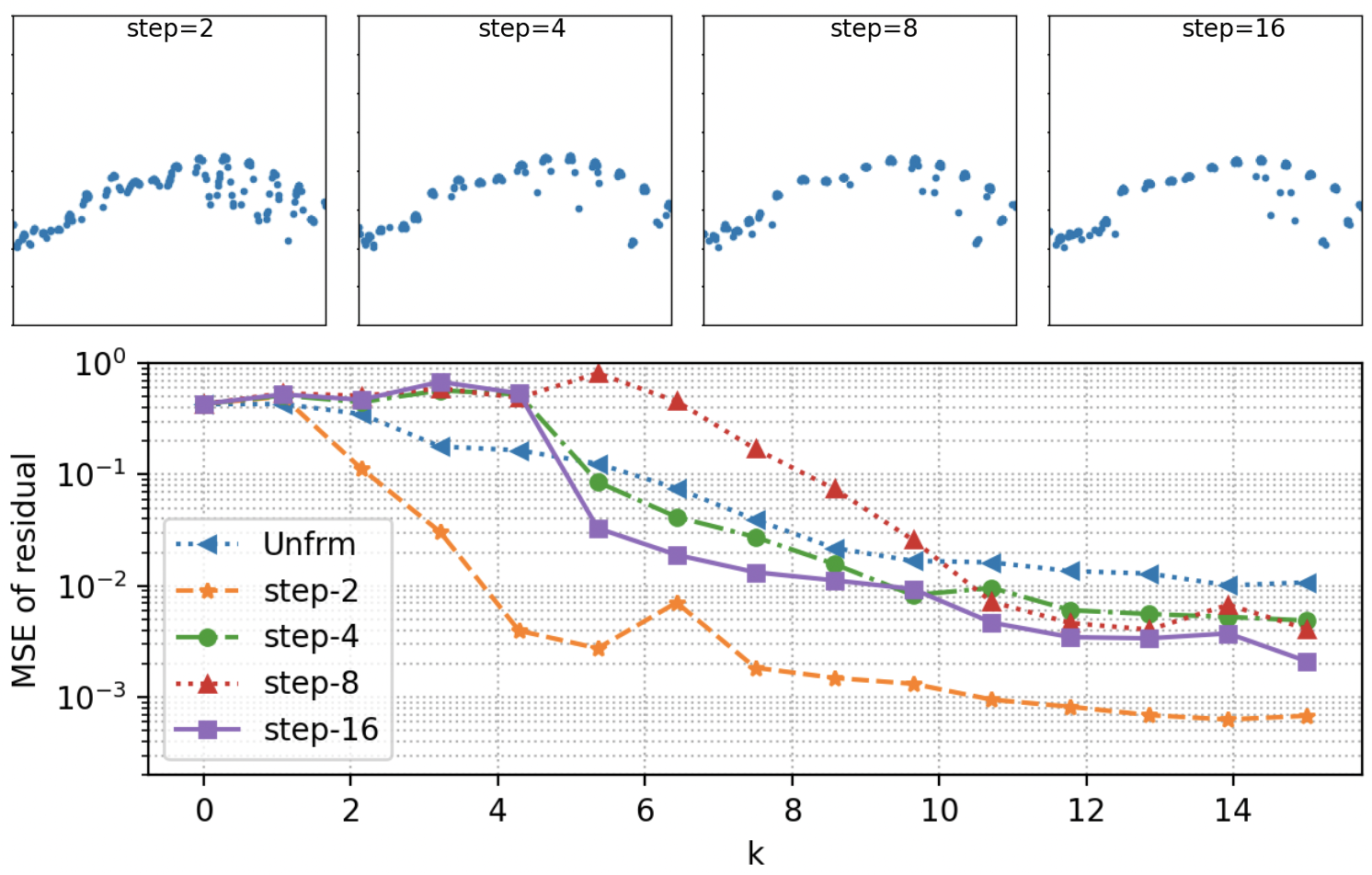}
	\caption{WbAR for the multi-scale problem with different adversarial attack steps ($T$) at $k=0$.}
	\label{fig:step}
    \end{subfigure}
	\hfill
    \begin{subfigure}[b]{.45\textwidth}
    	\centering
        \includegraphics[width=\textwidth]{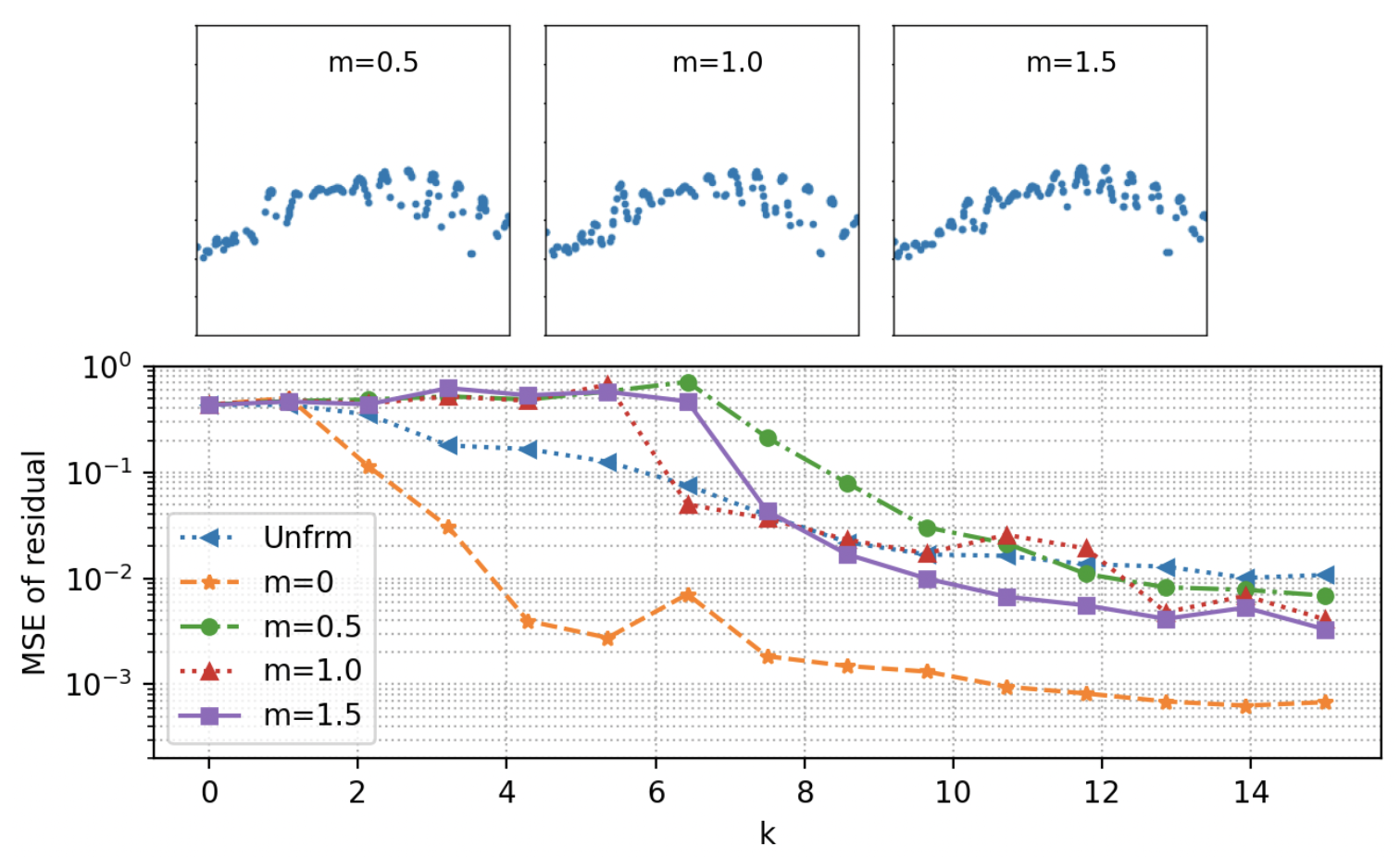}
	\caption{WbAR for the multi-scale problem with different adversarial attack revisiting ($m$) at $k=0$.}
	\label{fig:m}
    \end{subfigure}
    
    \begin{subfigure}[b]{.45\textwidth}
    	\centering
        \includegraphics[width=\textwidth]{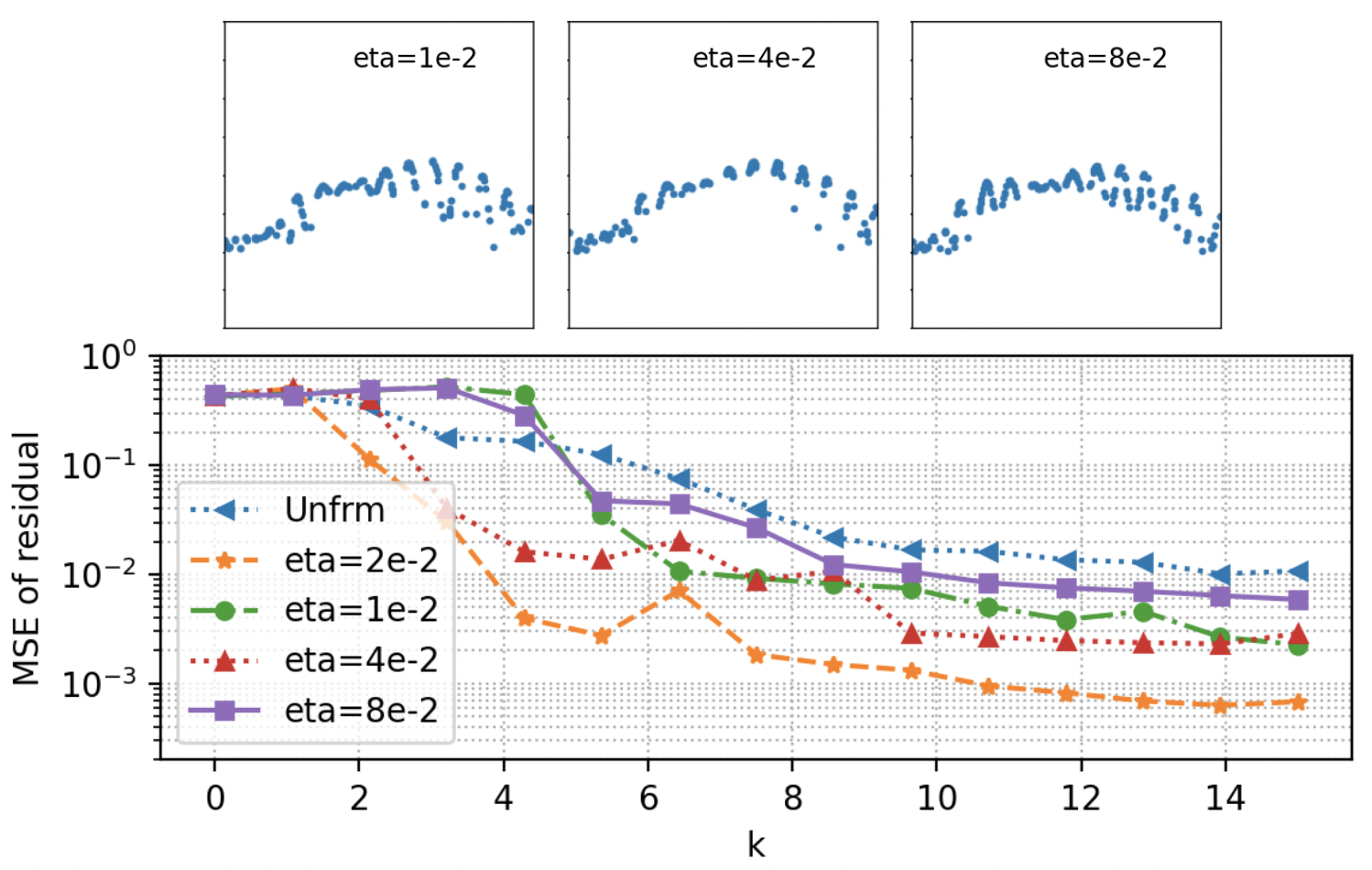}
	\caption{WbAR for the multi-scale problem with different adversarial attack step size ($\eta$) at $k=0$.}
	\label{fig:eta}
    \end{subfigure}
    \hfill
    \begin{subfigure}[b]{.45\textwidth}
    	\centering
        \includegraphics[width=\textwidth]{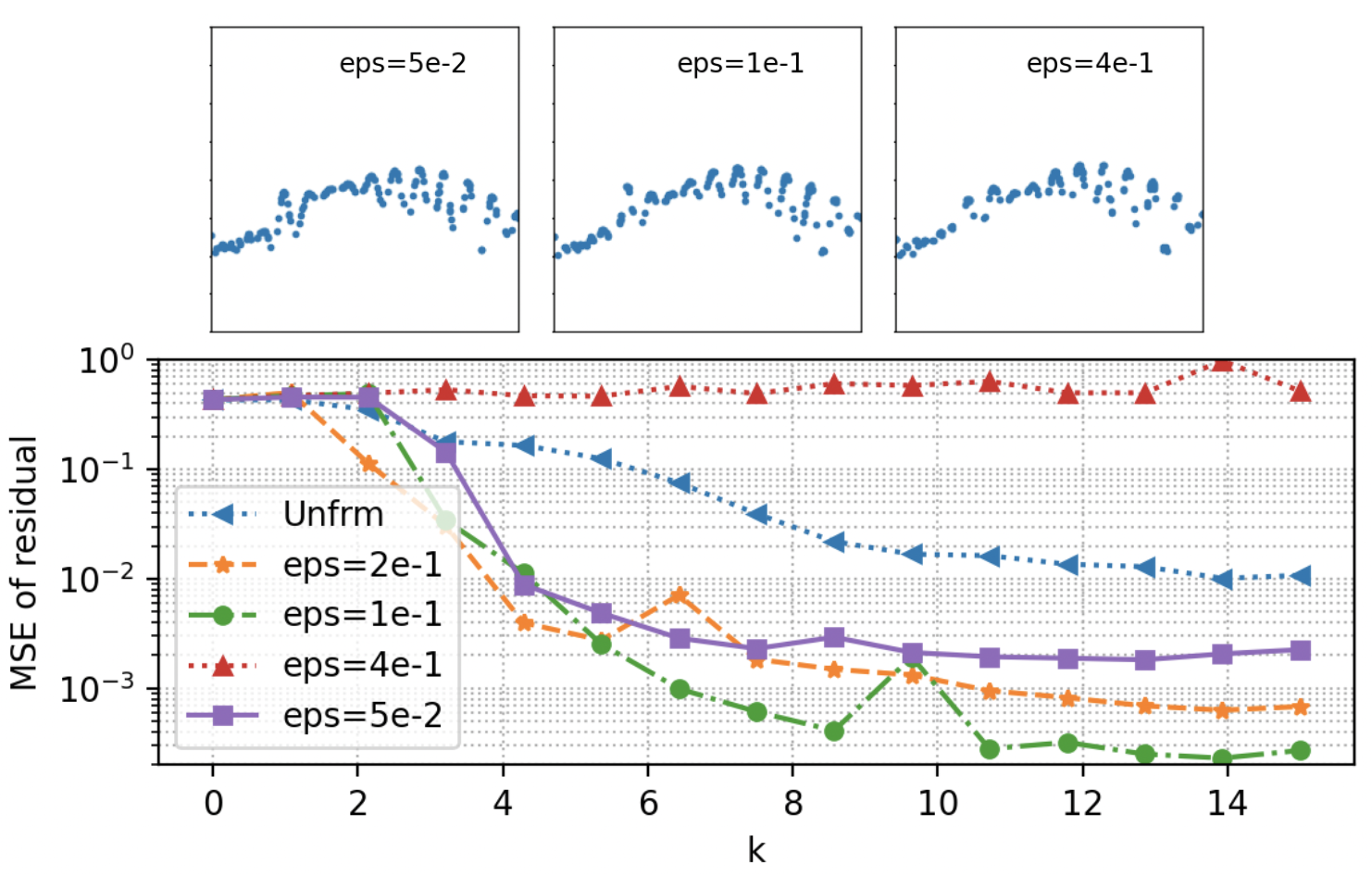}
	\caption{WbAR for the multi-scale problem with different adversarial attack maximum perturbation threshold ($\epsilon$) at $k=0$.}
	\label{fig:eps}
    \end{subfigure}
    \caption{Hyper-parameter studies for WbAR}
    \label{fig:hyper_parameter}
\end{figure*}

PGD is an algorithm that searches for local maxima iteratively, so more iteration steps ($T$) can bring higher accuracy. 
In adversarial training, more adversarial attack steps can reduce the distance of samples to local maximums of the residual. 
As shown in Figure \ref{fig:step}, when $T$ increases, the adversarial samples are more clustered to the local maximum. 
In the 1-dimensional multi-scale problem, we chose a much smaller $T$ to get better observations for each failure region rather than the maximum point only. 
Comparing different number of steps, $T=2$ has the best performance, while all tested adversarial training steps achieve better performance than the uniform sampling.

The revisiting mechanism selects adversarial samples of the most highest residuals. 
Thus, some small local maximums will be ignored if a larger $m$ is applied. 
The spread of adversarial samples will shrinks, because only large local maximums will be considered. 
As shown in Figure \ref{fig:m}, when $m$ increases, the adversarial samples have larger residual in average. 
In the multi-scale problem, we set $m=0$ to make sure no local maximums are ignored, because there are many small residual local maximums in the multi-scale problem. 
Comparing different revisiting lengths, $m=0$ has the best performance while all tested adversarial training steps achieve better performance than the uniform sampling.

The adversarial attack step size $\eta$ determines the smoothness of the iteration process. Large step size $\eta$ can make the the maximum search faster but may lead to oscillations in the search trajectory.
One can adjust $\eta$ according to the volume of the problem domain to adjust the searching speed. 
In Figure \ref{fig:eta}, when $\eta=$4e-2, PGD demonstrates better performance in locating the local maximum. 
However, we hope to sample on the entire failure regions instead of at the local maximum points only. 
Thus, $\eta=$2e-2 is applied to solve the equation and achieved good performance.

The maximum perturbation threshold $\epsilon$ determines the searching range. 
As shown in Figure \ref{fig:eps}, if we choose a smaller searching range, i.e. $\epsilon=$1e-1, better performance can be obtained. 
If we chose $\epsilon=$4e-1, the WbAR can fail the equation approximation of the multi-scale problem since the adversarial attack may reach a local maximum far away from the initial position. The maximum perturbation threshold $\epsilon$ should be selected based on the volume of the problem domain. In our experiments, usually we set  $\epsilon$ equals to one-tenth of the range of $x$. In the causal inferencing problems, we set $\epsilon$ equals to two-tenths of the range of $t$ to accelerate the causal inference speed. A lower $\epsilon$ can increase the accuracy for causal inferencing problems but decrease the training speed.

\subsection{Model Generalization} \label{sec:generalization}
The core concept of WbAR involves employing adversarial attack techniques to adaptively sample points from failure regions during neural network training. While initially developed for vanilla PINNs, this approach can be extended to numerous PINN variants, including those specifically designed for multiscale and high-dimensional problems.

For multiscale problems, Dubbed FMPINN (\cite{li2024solving}) represents a notable advancement for solving a particular class of multiscale elliptic PDEs. The FMPINN architecture effectively addresses multiscale challenges by combining Fourier feature embeddings with a mixed formulation of physics-informed constraints. Consider the following type of multiscale problem:
\begin{align}
&-\text{div}(A^\epsilon(x)\nabla u^\epsilon(x))=f(x), x\in \Omega, \notag \\
&\mathcal{B}u^\epsilon(x) = g(x), x\in \partial \Omega
\end{align}

Under the setting of Example 4 in \cite{li2024solving}, we consider a two-dimensional boundary value problem with Dirichlet conditions on the regular domain $\Omega = [-1,1] \times [-1,1]$. The forcing function is given by $f(x_1,x_2) = 1$, and the multi-frequency coeﬃcient is:
\begin{align*}
A^\epsilon(x_1,x_2) = \prod_{i=1}^5 &\Big((1+0.5\cos(2^i\pi(x_1+x_2))) \cdot \\
&(1+0.5\sin(2^i\pi(x_2-3x_1)))\Big)
\end{align*}

\begin{figure}[hbt]
\centering
\includegraphics[width=.45\textwidth]{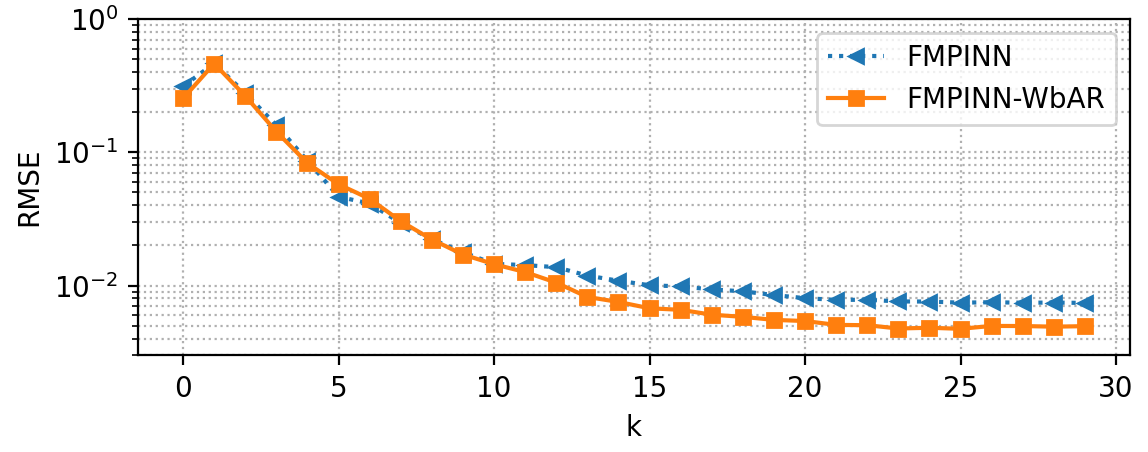}
\caption{Testing results for FMPINN and FMPINN-WbAR.}
\label{fig:fmpinn}
\end{figure}

The original implementation from the GitHub repository was used to reproduce the  example. The training scheme of FMPINN employs $5,000$ collocation points per batch across $50,000$ epochs, with resampling performed at every epoch. To adapt WbAR to this framework, adversarial samples are computed every $5,000$ epochs starting from $10,000$ epochs onward. For each subsequent epoch, the batch consists of $2,500$ resampled collocation points mixed with $2,500$ pre-calculated adversarial samples. As illustrated in Figure \ref{fig:fmpinn}, WbAR-aided sampling achieves a lower RMSE upon convergence compared to the original sampling scheme.

For high-dimensional problems, Stochastic Dimension Gradient Descent (SDGD) (\cite{hu2024tackling}) achieves efficient performance by
decomposing the gradients of the PDE and PINN residuals into dimension-specific components. During each training iteration, it randomly samples a subset of these dimensional gradients to update the model. Consider a Poisson equation defined in the unit ball $\mathbb{B}^d$ with a zero boundary condition:
\begin{align}
&\Delta u(x) = g(x), x\in \mathbb{B}^d \notag \\
&g(x) = \Delta u_{exact}(x)
\end{align}
The exact solution is
\begin{align}
u_{exact}(x) = &(1-\|x\|_2^2) \cdot \notag \\
&\Big(\sum_{i=1}^{d-1}c_i \sin(x_i+ \cos(x_{i+1})+x_{i+1}\cos(x_i))\Big) \notag
\end{align}
where $c_i\sim \mathcal{N}(0,1)$.

We utilized the implementation from the authors' GitHub repository for reproducibility. In our tests, we set the dimension $d=10$ and the PINN model depth $PINN_L=10$. To adapt WbAR, adversarial samples are computed every $500$ epochs, starting from epoch $500$ onward. Each training batch consists of $50$ resampled collocation points mixed with $50$ precomputed adversarial samples. As shown in Figure, WbAR achieves a lower RMSE under this configuration.

\begin{figure}[hbt]
\centering
\includegraphics[width=.42\textwidth]{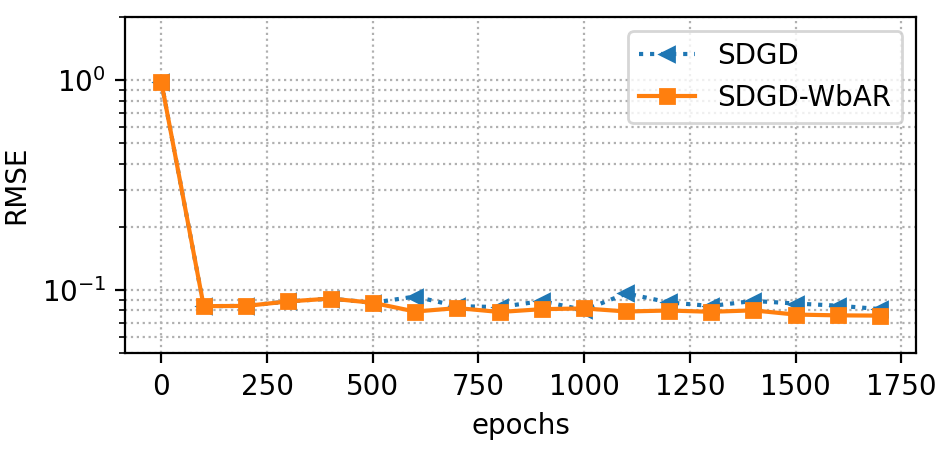}
\caption{Testing results for SDGD and SDGD-WbAR.}
\label{fig:sdgd}
\end{figure}

\section{Discussion}

\textit{WbAR v.s. SAIS}: SAIS is designed to be one of the best sampling algorithms for problems with Gaussian-shaped residual distributions. Although a mixture-of-Gaussian extension has been proposed, the residual distribution still needs to be known. In contrast, WbAR is designed to adapt to more complex residual distributions. Its performance may be inferior to SAIS in problems where the residual is Gaussian or follows other simple distributions, which are not presented in this script. Therefore, one should carefully choose between WbAR and SAIS based on the residual distribution.

\textit{WbAR v.s. RAR}: Although the two sampling algorithms are very similar, there are several key differences. First, WbAR is based on random walk and gradient descent, preferring to find all the local maxima of the residual. In contrast, RAR searches for the global maximum using Monte Carlo methods and tends to concentrate most of the samples around the global maximum, especially when a large number of Monte Carlo samples are generated. Depending on how many samples are chosen, the training details, and the specific problems, there is still insufficient evidence to conclude whether finding local maximums is better or not.  Second, in high-dimensional scenarios, the efficiency of Monte Carlo searching decreases, making gradient-based optimal searching algorithms more advantageous.

\textit{The computational cost}: Although WbAR appears to generate adversarial samples in one run, the adversarial attack requires several gradient descent steps, which are computationally costly. Moreover, unlike RAR, which only infers the model, WbAR computes gradients using auto-grad, resulting in significantly higher memory and computational costs. However, as shown in Table \ref{tab:computation_cost}, for high-dimensional problems, RAR requires significantly more GPU memory to achieve comparable performance with WbAR. In the context of PINN training, the computational overhead of WbAR, SAIS, and RAR sampling algorithms remains negligible.

\begin{figure}[hbt]
\centering
\includegraphics[width=.4\textwidth]{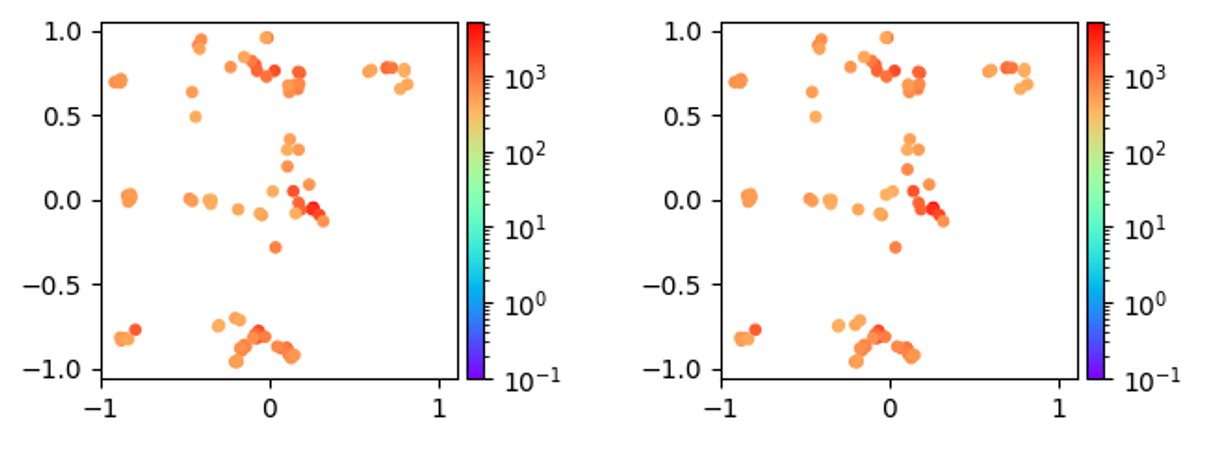}
\caption{WbAR samples for the 2d Poisson equation at $k=1$ (left: step-10, right: step-20).}
\label{fig:discussion_params}
\end{figure}

\begin{table}[htb]
\small
\caption{Computational cost (RTX 5090 32GB, 25 vCPU Intel(R) Xeon(R) Platinum 8470Q).}
\label{tab:computation_cost}
\centering
\scriptsize
\begin{tabular}{l   l r r } 
&\multicolumn{3}{c}{High-dimensional Poisson} \\
\hline
& algorithm & time (sec.) &  GPU memory (MB)\\
\hline
\multirow{7}{*}{sampling time} 
 & SAIS & 0.744 & 71.79\\
 & RAR-150 & 0.801 & 10334.87\\
 & RAR-80 & 0.782 & 5520.07\\
 & RAR-40 & 0.742 & 2771.09\\
 & RAR-2 & 0.704 & 153.89\\
 & FIPINN-R-1 & 249.0 & 616.77\\
 & WbAR & 1.842 & 773.04\\
\hline
training time & - & 11628 & -  \\
\hline
\\
&\multicolumn{3}{c}{Two-dimensional Poisson} \\
\hline
& algorithm & time (sec.) &  GPU memory (MB)\\
\hline
\multirow{7}{*}{sampling time} 
 & SAIS & 0.541 & 17.14\\
 & RAR-15 & 0.504 & 30.08\\
 & RAR-10 & 0.506 & 25.50\\
 & RAR-2 & 0.478 & 18.16\\
 & FIPINN-R-1 & 7.477 & 29.91\\
 & WbAR & 0.703 & 22.82\\
\hline
training time & - & 1244 & -  \\
\hline
\end{tabular}
\end{table}

\textit{Hyper-parameter selection}: Most of the hyper-parameters are inherited from the original adversarial attack algorithm. Although we discussed the effects of the hyper-parameters of WbAR, there are still several tips for hyper-parameter selection. WbAR attempts to find all the local maxima of the residual, so examining the residual map and the WbAR samples can provide useful insights. First, the maximum residual of WbAR samples should be close to the maximum residual of the problem domain. Second, the distribution of WbAR samples should coincide with the residual distribution, leaving low residual regions blank. More iteration steps and a smaller step size will ensure the first tip, while a larger attack threshold will ensure the second.

In the above experiments, we kept $\epsilon=2e-1$, $\eta=2e-2$, and $step=20$ for most cases to ensure consistency. However, if the gradient is not complicated, steps beyond 10 are not particularly useful since  $10\times\eta=\epsilon$. As shown in Figure \ref{fig:discussion_params}, samples generated with step-10 and step-20 are almost identical. In fact, step-10 achieved better results in our tests. The model trained with step-10 achieved an RMSE of 1.7 after the first training iteration, whereas the model trained with step-20 reached an RMSE of 2.0 after the first iteration. This phenomenon is consistent with the observations while training with RAR-$x$, i.e., concentrating samples too much can decrease model performance. This may be because the samples disrupt the balance or fail to accurately describe the residual distribution.

\textit{Future works}: We found that WbAR can also be applied in adaptive boundary points generation, which is one of our future works. Iterative training increases the number of samples, significantly raising the training cost as multiple iterations are required. We aim to explore which samples actually support the model weights through another adversarial attack process and reduce the redundant samples.

\section{Conclusion}

In this work, we propose the WbAR scheme which locates failure regions of PINNs through the PGD-based adversarial attack. 
The adversarial attack can adaptively generate adversarial samples on failure regions despite their number, size, distribution or dimension. 
As a result, WbAR can effectively locate and shrink failure regions when solving complex PDEs, especially those involving multi-scale behaviors or solutions with sharp or oscillatory characteristics. 
The effectiveness of the proposed WbAR has been verified with numerical experiments including an elliptic equation with multi-scale coefficients, Poisson equations with multi-peak solutions, and Burgers’ equation with sharp solutions. 
Numerical experiments show that WbAR is self-adapt to various problems benefiting from characteristics of the adversarial attack. As a result, WbAR has the potential to serve as a general tool for solving complex PDEs in the framework of PINNs.

In the WbAR scheme, there are various adversarial attack methods, including non-gradient based white-box attack methods and black-box attack methods, that may be particular effective to particular type of problems. 
For the adversarial attack, we chose a target attack method which maximize the residual to locate failure regions. 
Other adversarial attack mechanisms such as non-target attacks can be explored in further works. 
The iterative training framework can also be further optimized to accelerate the training process, such as adding adversarial samples during training, etc. 
Beyond the adversarial attack scheme, the failure region locating is actually a non-convex optimization problem which finds all the local maxima. 
A broader optimization based approaches can be explored for failure region searching. Moreover, adversarial samples during the iterative training are Markov given the residual map, while other sampling strategies assume the samplings are independent. 
The behavior of different sampling strategies can also be discussed under the stochastic process.

\section{Acknowledgement}
This document is the results of the research project funded by National Natural Science Foundation of China U21B2075; the National Key R \& D Program of China 2023YFC2205900, 2023YFC2205903; National Natural Science Foundation of China 12371419, 12171123, 12271130; the Natural Sciences Foundation of Heilongjiang Province ZD2022A001.

\bibliographystyle{cas-model2-names}

\bibliography{references.bib}

\end{document}